\documentclass{article}
\PassOptionsToPackage{varqu}{inconsolata}
\usepackage{colm2024_conference}

\usepackage[utf8]{inputenc} 
\usepackage[T1]{fontenc}    
\usepackage{booktabs}       
\usepackage{amsfonts}       
\usepackage{amssymb}        
\usepackage{amsmath}        
\usepackage{marvosym}
\usepackage{fontawesome}
\usepackage{graphicx}       
\usepackage{caption}        
\usepackage{colortbl}
\usepackage{array}
\usepackage{tabularx}
\usepackage{enumitem}       
\usepackage{multirow}       
\usepackage{rotating}       
\usepackage{algorithm}      
\usepackage{algpseudocode}  
\usepackage{nicefrac}       
\usepackage{adjustbox}      
\usepackage{tcolorbox}
\tcbuselibrary{breakable,skins}

\newtcolorbox{takeaway}{
  enhanced,
  boxrule=0pt,
  frame hidden,
  borderline west={3pt}{0pt}{red!70!black},
  colback=red!5,
  sharp corners,
  left=10pt, right=10pt, top=6pt, bottom=6pt,
}
\newtcolorbox{finding}{
  enhanced,
  boxrule=0pt,
  frame hidden,
  borderline west={3pt}{0pt}{blue!70!black},
  colback=blue!4,
  sharp corners,
  left=10pt, right=10pt, top=6pt, bottom=6pt,
}
\newtcolorbox{examplebox}{
  enhanced,
  boxrule=0pt,
  frame hidden,
  borderline west={3pt}{0pt}{green!55!black},
  colback=green!5,
  sharp corners,
  left=10pt, right=10pt, top=6pt, bottom=6pt,
}
\definecolor{boxbrown}{HTML}{862D1A}
\definecolor{boxwarm}{HTML}{F7F4ED}
\newtcolorbox{evalcontextbox}{
  enhanced,
  colback=boxwarm!30,
  colframe=boxbrown!80,
  colbacktitle=boxwarm,
  coltitle=boxbrown!90!black,
  fonttitle=\small,
  title={\textbf{Example: Prompt-Specific Evaluation Context}},
  boxrule=0.5pt,
  arc=1.5pt,
  left=8pt, right=8pt, top=4pt, bottom=4pt,
  toptitle=3pt, bottomtitle=3pt,
}
\newtcolorbox{insight}{
  enhanced,
  boxrule=0pt,
  frame hidden,
  borderline west={3pt}{0pt}{violet!65!black},
  colback=violet!5,
  sharp corners,
  left=10pt, right=10pt, top=6pt, bottom=6pt,
}
\tcbset{
  appboxbase/.style={
    enhanced,
    breakable,
    boxrule=0pt,
    frame hidden,
    sharp corners,
    left=10pt, right=10pt, bottom=6pt,
    before upper={\small\raggedright\sloppy},
    fonttitle=\small\bfseries,
    coltitle=black,
    colbacktitle=white,
    boxed title style={frame hidden, boxrule=0pt, colback=white, sharp corners},
    attach boxed title to top left={xshift=8pt, yshift=-2pt},
    top=12pt,
  },
  appinputbox/.style={
    appboxbase,
    borderline west={3pt}{0pt}{blue!70!black},
    colback=blue!4,
  },
  appoutputbox/.style={
    appboxbase,
    borderline west={3pt}{0pt}{green!55!black},
    colback=green!5,
  },
  appevalbox/.style={
    appboxbase,
    borderline west={3pt}{0pt}{orange!80!black},
    colback=orange!5,
  },
  appscorebox/.style={
    appboxbase,
    borderline west={3pt}{0pt}{red!70!black},
    colback=red!4,
  },
}
\usepackage{listings}
\lstdefinestyle{appendixjson}{
  basicstyle=\ttfamily\tiny,
  breaklines=true,
  breakatwhitespace=false,
  columns=fullflexible,
  showstringspaces=false,
  frame=single,
  framerule=0.2pt,
  xleftmargin=0pt,
  xrightmargin=0pt
}
\usepackage{amsthm}

\newtcolorbox{defbox}[1][]{
  enhanced,
  boxrule=0pt,
  frame hidden,
  borderline west={3pt}{0pt}{headerblue!60},
  colback=headerblue!3,
  sharp corners,
  left=10pt, right=10pt, top=6pt, bottom=6pt,
}
\newcounter{definitionctr}
\renewcommand{\thedefinitionctr}{\arabic{definitionctr}}

\newcommand{\bench}{\textsc{SearchGen-Bench}}
\newcommand{\dataset}{\textsc{SearchGen-20K}}
\newcommand{\searchcorpus}{\textsc{SearchGen-Corpus-1M}}
\newcommand{\CondNS}{\textsc{No Search}}
\newcommand{\CondGS}{\textsc{Blind Search}}
\newcommand{\CondGAS}{\textsc{Generator-Adaptive Search}}
\newcommand{\website}{https://haozheh3.github.io/SearchGen}

\definecolor{headerblue}{RGB}{41,65,122}
\definecolor{rowgray}{RGB}{245,247,250}
\definecolor{grouptop}{RGB}{230,240,255}
\hypersetup{
  colorlinks=true,
  linkcolor=headerblue,
  citecolor=headerblue,
  urlcolor=headerblue,
}
\newcolumntype{R}{>{\raggedleft\arraybackslash}p{1.05cm}}

\title{Search Beyond What Can Be Taught: 
Evolving the Knowledge Boundary in Agentic Visual Generation}

\author{
    \textbf{Haozhe Wang}$^{1}$ \
    \textbf{Weijia Feng}$^{3}$ \
    \textbf{Jinpeng Yu}$^{3}$ \
    \textbf{Che Liu}$^{4}$ \
    \textbf{Ping Nie}$^{2}$ \
    \textbf{Fangzhen Lin}$^{1}$ \
    \textbf{Jiaming Liu}$^{3,\text{\Letter}}$ \
    \textbf{Ruihua Huang}$^{3}$ \
    \textbf{Jimmy Lin}$^{2}$ \
    \textbf{Wenhu Chen}$^{2}$ \
    \textbf{Cong Wei}$^{2,\text{\Letter}}$ \vspace{0.3em} \\
    $^{1}$Hong Kong University of Science and Technology \
    $^{2}$University of Waterloo \\
    $^{3}$Qwen Applications \
    $^{4}$Imperial College London
}

\begin{document}

{
  \renewcommand{\thefootnote}{} 
  \footnotetext{\Letter: Corresponding Authors.}
}

\maketitle

\begin{abstract}
    Visual generators excel at rendering, but they confidently fabricate what they do not know.
    User requests are unbounded, evolving, and deeply long-tailed: new characters, trending entities, post-cutoff events, etc.
    This \emph{world-knowledge bottleneck} is structural: generators are trained on fixed corpora,
    but the visual world is open-ended.
    We construct \textbf{\dataset{}} and \textbf{\bench{}}, 20{,}939 prompt records spanning twelve failure categories
    and twenty-two domains, paired with a pre-executed multimodal \textbf{\searchcorpus{}} to facilitate offline, reproducible research.
    On \bench{}, frontier open generators score only 21--28 out of 100, a 40-point collapse
    invisible to existing benchmarks. The natural remedy to this knowledge bottleneck is to employ search tools, enabling \emph{agentic visual generation}. But we found that naive search fails: it retrieves indiscriminately,
    injecting noise into prompts the generator already handles.
     We trace the root cause to a \emph{generator-specific{, evolving} knowledge boundary} --- the divide
  between what a generator can internalize through training and what must remain in
  external context --- and show that this boundary, though hard to specify in advance,
  is discoverable through a teach-then-search co-training framework.
  Even a minimal recipe of this co-training produces monotonic improvement, laying the foundation for recursive self-improvement of visual generation that meets world-knowledge grounded requests.
\textbf{  We release code, model and the full dataset, co-training corpus, and search corpus} as a
  replayable {harness} for {tool-augmented,} world-knowledge-grounded visual generation.
\end{abstract}

\begin{figure}[!h]
\centering
\vspace{-.1cm}
\includegraphics[width=\textwidth]{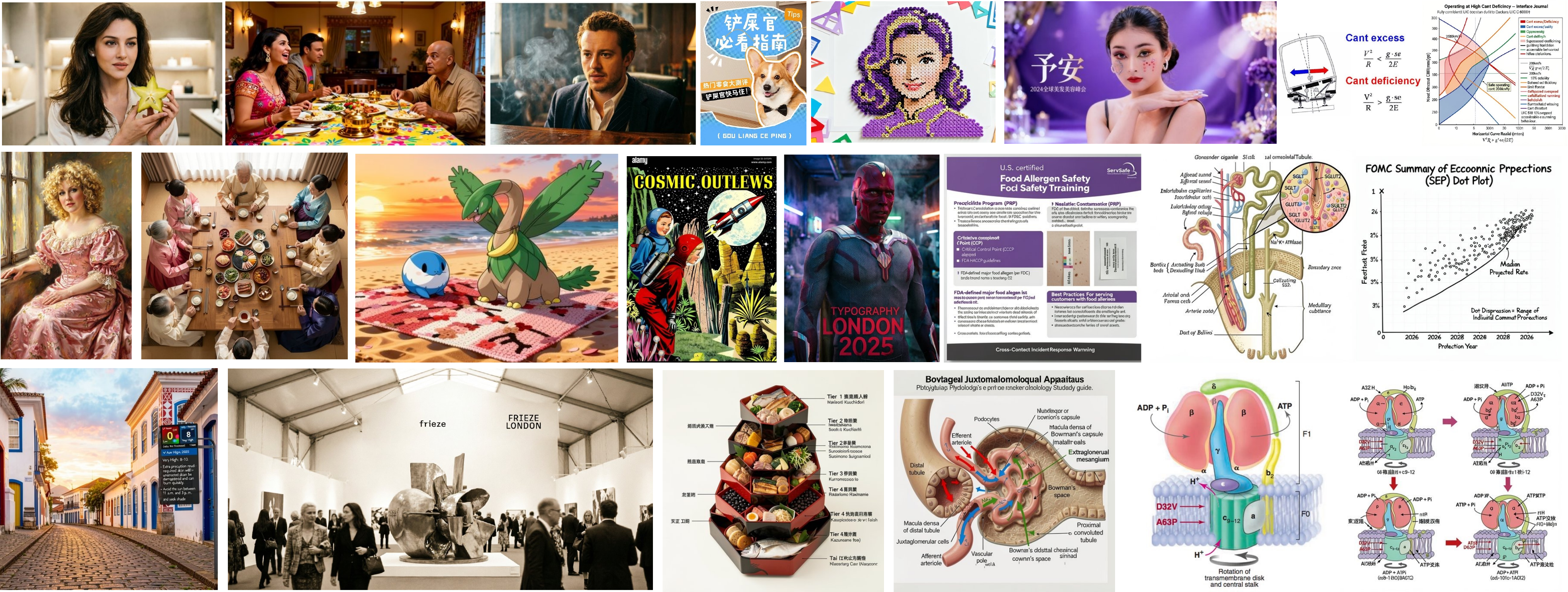}

\caption{\small\textbf{Representative search-augmented generations from \dataset{}, spanning all twelve failure categories identified from 20,840 production prompts.} \dataset{} captures the production-scale diverse user requests that demand the unbounded, evolving, and deeply long-tailed world knowledge. }
\label{fig:user_diversity}
    \vspace{-0.3em}
\end{figure}

\begin{figure}[t]
\centering
\includegraphics[width=\textwidth]{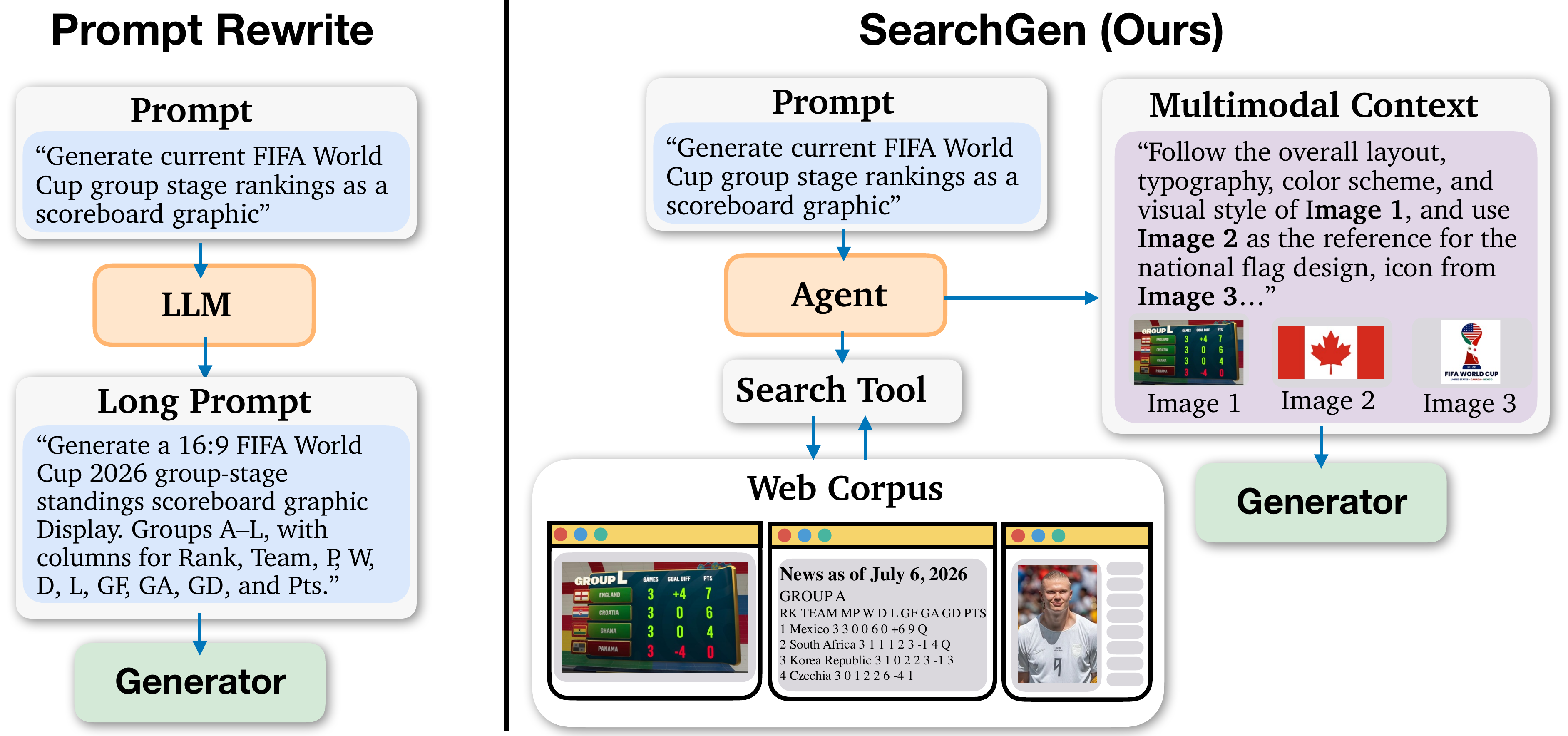}
\caption{\textbf{Two paradigms for visual generation.}
(\textit{Left}) Prompt rewriting relies on an LLM to expand the user query into a longer textual prompt, which is then passed directly to the generator.
(\textit{Right}) In contrast, our approach equips an agent with a search tool to retrieve relevant knowledge and visual references from a web corpus; the agent then organizes the retrieved evidence into multimodal context and provides it to the generator.}
\label{fig:teaser}
\vspace{-.3cm}
\end{figure}

\section{Introduction}
\label{sec:intro}

Ask a frontier image generator for the mascot of the 2025 Osaka Expo: you get a polished, confident fabrication.
Ask for a historically accurate Spartan phalanx: you get anachronistic armor rendered in exquisite detail.
Modern generators produce complex scenes with precise lighting and coherent structure, saturating on standard benchmarks~\cite{genaibench,blackforestlabs2025flux,esser2024scaling}, yet they still fail on a substantial class of real-world user requests (Figure~\ref{fig:user_diversity}).
These failures reflect a \emph{world-knowledge} bottleneck rather than a visual-synthesis bottleneck.
Visual generators are trained on fixed corpora with inherent knowledge cutoffs, whereas user requests are \emph{unbounded, evolving, and deeply long-tailed}: new characters, regional cultural symbols, niche typography, historical artifacts, and recent events.
The world knowledge bottleneck motivates external knowledge access as a natural complement to visual generation, analogous to illustrators consulting references before depicting unfamiliar concepts.

To systematically study this problem, we analyze over 20{,}000 user prompts from production-level AIGC platforms~\citep{qwenimage} and identify twelve major failure modes.
Based on these requests and failure modes, we construct \dataset{}, a collection of world-knowledge-grounded prompts spanning twenty-two domains and diverse global cultures, each annotated with fine-grained visual checklists for fully automated assessment (\S\ref{sec:benchmark}).
On \bench{}, frontier open generators achieve only $21$--$28$ out of $100$, at most half of the commercial API ceiling.
These results reveal a substantial evaluation gap: existing benchmarks largely fail to capture the live and dynamic world-knowledge failures in visual generation.
\begin{center}
    \textbf{Can search guide generators as references guide illustrators?} 
\end{center}

{This renders an agentic visual generation problem: a generator is equipped with an agentic reasoner that decides when and how to utilize search tools for knowledge acquisition.}
Through extensive experiments, our results show that naive search fails: agentic reasoner that triggers a search blindly for every prompt degrades the quality of a substantial fraction of prompts.
Why does naive search fail? We trace the root cause to two distinct issues that must be solved jointly.

The first is when to search. The world knowledge required for visual generation splits along a structural axis into what generators can internalize and what must remain in context. Some knowledge is stable and low-dimensional: a character's canonical appearance, a flag's fixed geometry. Once encountered through augmented supervision, such knowledge migrates into generator parameters and search becomes superfluous. Other knowledge is inherently contextual: it evolves faster than retraining cycles, sits too deep in the long tail for reliable learning~\cite{valiant1984theory}, or requires per-instance compositional reasoning that resists parameterization. The extreme long tail of visual entities in real requests (Figure~\ref{fig:entity_longtail}) confirms that this contextual stratum is large. This creates a knowledge boundary between the two regimes, and this boundary is generator-specific {and evolving}: it shifts as the generator improves, so a search policy that helps a weaker generator can harm a stronger one.

The second is what search returns. Even when search is correctly triggered, every result injects noise into the generator's input space: irrelevant visual detail, stylistic contamination, spurious structure. The open-weight generators we tested treat all conditioning signals as authoritative, with no capacity to distinguish useful reference content from noise.

\begin{figure}[t]
\centering
\includegraphics[width=0.85\textwidth]{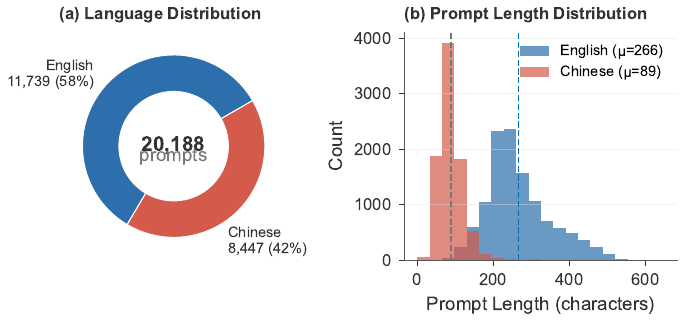}
\caption{\small\textbf{Bilingual composition and prompt length distribution of \dataset{}.}
Left: English (58\%) and Chinese (42\%) proportions. Right: bimodal prompt length
distribution -- Chinese prompts are concise (mean 89 characters) while English prompts
are more elaborate (mean 266 characters), reflecting authentic cross-lingual user behavior
rather than translated templates.}
\label{fig:language_prompt_length}
\vspace{-.3cm}
\end{figure}
We address both issues through two interlocking components. First, we introduce a \emph{noise-resistant} agentic reasoner that builds what to render for generator: the reasoner executes a three-stage gate--filter--integrate pipeline that controls when to search, what retrieved content to retain, and how external information is integrated into generation (\S\ref{sec:atomic}). This pipeline builds noise resistance into the reasoner so that noise is suppressed before reaching the generator. Second, we propose to co-train the reasoner with the visual generator: online iterative DPO teaches the generator to internalize world-knowledge that it can absorb and build noise-resistance to imperfect search-augmented inputs, while rejection-sampling finetuning calibrates the search reasoner to search only for what the generator cannot be taught (\S\ref{sec:training}).  

On \bench{}, this single-iteration recipe produces monotonic gains from \CondNS{} through \CondGS{} to \CondGAS{}, exceeding frontier generators paired with frontier VLMs.
The recipe is deliberately minimal,  yet we envision this co-training framework lays the foundation for a recursive self-improvement flywheel in visual generation: the generator progressively expanding rendering knowledge while the reasoner progressively adjust its search scope, catering to production-scale text-to-image requests that demand dynamically evolving world knowledge. 

\textbf{Contributions.}
\begin{enumerate}[leftmargin=*,itemsep=2pt]
\item We introduce \dataset{} and \bench{}, a large-scale dataset of 20{,}939 world-knowledge-grounded text-to-image prompt records spanning twelve failure categories and twenty-two domains, together with a pre-executed multimodal \searchcorpus{} for reproducible and democratized research.
\item We show that search-intensive visual generation exposes a substantial evaluation gap: frontier open generators collapse by up to 40 points on \bench{}, a gap that shifts as generators improve and that current benchmarks render entirely invisible.
\item We show that naive search introduces \emph{search noise}, causing concept corruption and copy effects, and identify the underlying challenge as a searcher--generator coordination problem governed by a generator-specific knowledge boundary.
\item We propose a co-training framework with a noise-resistant agentic search protocol. We validate that the minimalist recipe, an 8B agentic reasoner with search tools, jointly calibrated with the generator outperforms both pure generation and undiscriminating search by better respecting the knowledge boundary.
\end{enumerate}

\section{The World-Knowledge Bottleneck}
\label{sec:benchmark}
Faithful generation is limited less by rendering than by \emph{knowledge}. A visual generator is trained on a fixed corpus, but user requests draw on a world that is open-ended, evolving, and long-tailed; the mismatch between the two is the \emph{world-knowledge bottleneck}. The bottleneck is compounded by a structural blind spot: generators are trained to always produce an image, so they have no mechanism for recognizing what they do not know, no way to say ``I lack the knowledge to render this faithfully,'' and no protocol for acquiring it. This section characterizes the bottleneck end to end: we map the landscape of world-knowledge-grounded generation from real user behavior, construct a large-scale dataset and benchmark, quantify how severely it degrades frontier generators, and show that naive search introduces new failures rather than resolving it.

\subsection{What Users Actually Ask For} 
\label{sec:taxonomy}

\begin{figure}[t]
\centering
\includegraphics[width=0.75\textwidth]{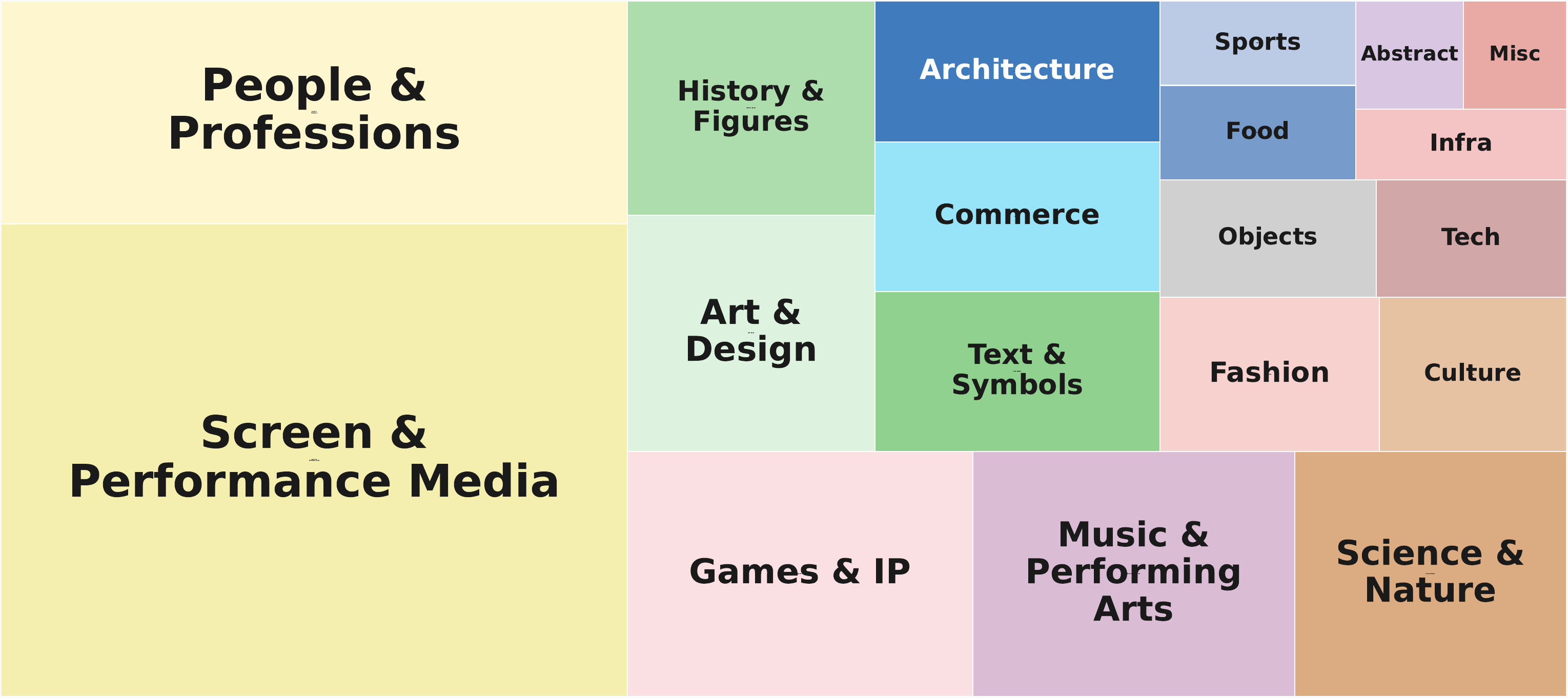}
\caption{\small\textbf{\dataset{} spans diverse, long-tailed domains.}
Treemap of benchmark mass across domain categories (area reflects relative prompt counts). The cross-category severity structure between failure modes and domains is deferred to Appendix~\ref{app:benchmark} (Figure~\ref{fig:heatmap_severity}), where uniform severity across all cells rules out the hypothesis that failures concentrate in a few niche categories.}
\label{fig:bench_overview}
\vspace{-.3cm}
\end{figure}

To ground the benchmark in real-user demand, we analyzed 20{,}840 prompts from a production text-to-image platform.
The landscape is far more diverse, long-tailed, and rapidly evolving than existing benchmarks suggest.
We discover twelve recurring failure categories distilled from real-world user requests (Table~\ref{tab:taxonomy}); Figure~\ref{fig:user_diversity} shows example generations representative of these failure modes.

As shown in Table~\ref{tab:taxonomy}, the twelve categories split along a structural axis that directly determines search design.
Some require \emph{visual} references that resist linguistic specification -- a game character's costume silhouette, a national flag's precise color fields -- that no textual description can fully specify but that a generator can readily condition on once retrieved. For example, generating \emph{Jingliu from Honkai: Star Rail (a game character)}  demands a precise visual identity (costume, weapon, color palette).

Other requests require \emph{textual} knowledge for which no visual shortcut exists or images alone cannot convey. For instance, \emph{a China dynastic timeline with dates and founders} relies entirely on structured factual content (exact names, years, and causal orderings), and \emph{an Aztec-style infographic of DNA replication} requires precise biochemical labels and process orderings for which visual shortcuts rarely exists but encyclopedic text can supply.

Many categories sit in between, requiring both modalities simultaneously: \emph{a traditional Oaxacan alebrije dragon} demands visual references for culturally specific chromatic conventions \emph{and} textual knowledge of what distinguishes an alebrije from a generic dragon; \emph{the mascot for the 2025 Osaka Expo} requires temporally fresh visual references of the character \emph{and} textual context that postdates model training.

\begin{table}[t]
  \centering
  \caption{\textbf{Taxonomy of search-intensive visual generation.} Twelve categories organized by failure source, each with a representative example prompt. The \emph{Modality} column indicates the dominant knowledge type required.}
  \label{tab:taxonomy}
  \small
  \renewcommand{\arraystretch}{1.1}
  \begin{tabularx}{\linewidth}{@{}lcX@{}}
  \toprule
  \textbf{Category} & \textbf{Modality} & \textbf{Example Prompt} \\
  \midrule
  \textbf{Temporal -- Recent}    & Both    & ``The mascot for the 2025 Osaka Expo in its official pose and colors'' \\
  \textbf{Temporal -- Current}   & Both & ``Current FIFA World Cup group stage rankings as a scoreboard graphic'' \\
  \textbf{Entity \& IP}          & Visual  & ``Jingliu from Honkai: Star Rail wielding her ice sword'' \\
  \textbf{Concept \& Symbol}     & Both & ``The national flag of Bhutan with the correct Druk dragon design'' \\
  \textbf{Factual \& Historical} & Both    & ``Spartan phalanx at Thermopylae with historically accurate bronze armor'' \\
  \textbf{Cultural Specificity}  & Both    & ``A traditional Oaxacan alebrije dragon with authentic chromatic patterns'' \\
  \textbf{Visual / UI / UX}      & Visual  & ``iOS 17 Weather app screenshot showing a thunderstorm animation'' \\
  \textbf{Data Visualization}    & Textual & ``China dynastic timeline with accurate dates and founding emperors'' \\
  \textbf{Text / Typography}     & Textual & ``Art Nouveau concert poster with period-authentic display lettering'' \\
  \textbf{Complex Composite}     & Both    & ``Aztec-style infographic of DNA replication with labeled process stages'' \\
  \textbf{Vague / Abstract}      & Textual & ``The feeling of nostalgia on a rainy afternoon in a small Japanese town'' \\
  \textbf{Implicit Reasoning}    & Both  & ``Cozy mountain cabin interior in the style of a Miyazaki film'' \\
  \bottomrule
  \end{tabularx}
\end{table}

\begin{figure}[t]
\centering
\includegraphics[width=\textwidth]{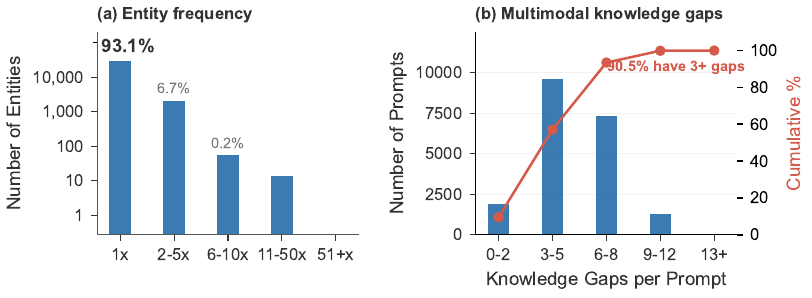}
\caption{\small\textbf{Dataset composition: entity long-tail and multimodal knowledge gaps.}
\textbf{(a)} Entity frequency distribution: 93.1\% of the 31{,}537 unique visual entities
appear in only a single prompt, confirming the extreme long-tailed nature of real-world
image generation requests.
\textbf{(b)} Multimodal knowledge gaps: prompts carry a mean of 5.2 simultaneous knowledge gaps
(reference slots + text knowledge slots + failure modes); 90.5\% carry three or more,
confirming the multi-constrained nature of the dataset.}
\label{fig:entity_longtail}
\vspace{-.3cm}
\end{figure}
This dual-modality structure, visual identity on one axis and factual precision on the other, directly motivates the modality-aware search protocol we introduce in \S\ref{sec:method}: neither visual nor textual search alone suffices, and the choice of modality must be made per knowledge gap. In this work, we use Google Search API with both image search and web-text search. {Image search and web search are thus two distinct tools the agentic reasoner utilizes between per knowledge gap (\S\ref{sec:atomic}).}

\subsection{Dataset Construction} 
\label{sec:bench_construct}

From the failure taxonomy (Table.~\ref{tab:taxonomy}), we construct \dataset{} spanning all twelve categories and twenty-two
domains. The dataset is intentionally bilingual (Figure~\ref{fig:language_prompt_length}): English prompts (58\%) average 266 characters and tend toward descriptive elaboration, while Chinese prompts (42\%) average 89 characters and favor dense, implicit specification, capturing authentic cross-lingual prompt behavior rather than translated templates. The full training dataset comprises 20{,}188 rows (20{,}187 unique prompts) with an average of 5.2
multimodal knowledge gaps per prompt (90.5\% carry three or more), annotated with
34{,}694 visual reference slots, 16{,}345 textual knowledge slots. 
We partition into train (20{,}188 rows with 128 held out for validation) and test (751) sets.

To construct this dataset, we first extract a seed database of 31{,}537 entities from production-scale user requests, each annotated with canonical names, estimated training-set frequency, ground-truth visual references (where available), and distinguishing attributes; the distribution is intentionally long-tailed, mirroring real user demand (Figure~\ref{fig:entity_longtail}a). With 93.1\% of entities appearing in only a single prompt, parametric learning from any feasible training dataset cannot cover this demand; search is structurally unavoidable for the tail.
From this seed database, we synthesize the \dataset{} dataset via human-brainstormed template-based generation and LLM-assisted rewriting.

During prompt synthesis, we deliberately employ an answer-first strategy. We ask a frontier model to select self-contained world-knowledge gaps based on the seed context. Therefore, each prompt carries \emph{knowledge gap references} by construction, structured slots specifying what visual or textual knowledge the generator is likely to lack, each with a severity label (critical, important, moderate or minimal) and a justification field. These annotations ease the burden of automated evaluation.

\begin{table}[t]
\centering
\caption{\textbf{We release \dataset{}, \searchcorpus{} and \bench{} which compose replayable environment for agentic visual generation.}
These assets provide agentic reasoning and generation traces collected across multiple reasoners and generators for the 20{,}188 training rows (20{,}187 unique prompt strings). \textbf{\searchcorpus{}} provides cached search sessions and downloads which emulate live web search offline, relieving the burden of costly search APIs.}
\label{tab:released_assets}
\footnotesize
\setlength{\tabcolsep}{4pt}
\begin{tabularx}{\columnwidth}{@{}lrX@{}}
\toprule
\textbf{Asset} & \textbf{Scale} & \textbf{Enables} \\
\midrule
\multicolumn{3}{@{}l}{\textit{\textbf{Training Trajectories} (20{,}188 rows)}} \\
Reasoning traces across various frontier reasoners     & 96{,}848  & search-policy / reward learning \\
Image Generations using different reasoning trace x generators          & 283{,}493 & preference \& distillation data \\
\midrule
\multicolumn{3}{@{}l}{\textit{\textbf{Search corpus} (archived, indexed)}} \\
Search sessions             & 159{,}027 & reproducible retrieval, no live API \\
\quad image / web           & 120{,}828 / 38{,}199 & \\
SERP hits (unique URLs)     & 1{,}097{,}961 & retrieval / grounding studies \\
Cached downloads            & 705{,}838 & fully offline replay \\
\quad image / web           & 599{,}859 / 105{,}979 & \\
\bottomrule
\end{tabularx}
\end{table}

\textbf{Released assets: a {replayable harness}.}
Reproducing search-augmented generation normally demands paid API access to both a search engine and a fleet of generators, and results drift as those services change. We remove that barrier. Alongside \dataset{}/\bench{}, we release the full training dataset: 96{,}848 reasoning traces collected from different frontier reasoners, and 283{,}493 image generations spanning multiple reasoner configurations and frontier generators (Table~\ref{tab:released_assets}). We further release a \searchcorpus{} with 159{,}027 archived image and web search sessions, 1{,}097{,}961 unique search URLs, and 705{,}838 cached downloads. Because every search is pre-executed and frozen, the pipeline can be replayed without live search API keys or search-result drift, turning an expensive workflow into an offline substrate for preference learning, reward modeling, search-policy design, and retrieval research. {Together, the gate--filter--integrate protocol, the co-training loop, and this frozen corpus constitute a reusable \emph{harness} for agentic, tool-augmented visual generation.}

\subsection{Evaluation Protocols}
\label{sec:eval_protocols}

Before we deliver our major empirical findings, we first briefly clarify our evaluation protocols for world-knowledge-grounded image generation. We aim to distinguish between world-knowledge failures and rendering failures with fully automated evaluation. Therefore, we employ two complementary set of evaluation score components. As Table~\ref{tab:eval_components} details, four components are \emph{knowledge-sensitive}: \emph{Checklist verification} tests 3--5 binary questions about key visual elements. \emph{Rubric-adaptive scoring} applies 2--4 category-tailored weighted dimensions.  \emph{Prompt faithfulness} measures overall adherence to the textual request, and \emph{visual reference fidelity} measures faithfulness to provided reference images (when applicable). These evaluation criteria tailor to prompt-specific knowledge demands.

The remaining five components are \emph{knowledge-invariant} dimensions that assess \emph{rendering quality} (bottom half of Table~\ref{tab:eval_components}): image quality, text rendering, AI naturalness, composition \& aesthetics, and physical plausibility. This decomposition comprehensively diagnoses the knowledge bottleneck and rendering gap. We provide an example below for concrete illustration. 

\begin{evalcontextbox}
\small
\textbf{Prompt:} \emph{``Paint Yang Chaoyue (a celebrity) on a 1970s Chinese rural field ridge; long-handled hoe; earth tones; 1970s film grain.''} \\[3pt]
\textbf{Checklist} (binary verification questions, each targeting one knowledge gap):
\begin{itemize}[leftmargin=1.2em,itemsep=1pt,topsep=1pt,parsep=0pt]
  \item Is the person recognizable as Yang Chaoyue (facial features and overall bearing)?
  \item Do the clothing and setting match a 1970s Chinese rural commune worker rather than modern workwear?
  \item Does the farm tool match the form of a traditional long-handled hoe?
  \item Does the image show the requested earth-tone palette and 1970s film-grain texture?
\end{itemize}
\vspace{1pt}
\textbf{Rubric} (category-specific weighted dimensions; the justification states what each dimension scores):
\begin{itemize}[leftmargin=1.2em,itemsep=1pt,topsep=1pt,parsep=0pt]
  \item \texttt{celebrity\_likeness}\,(0.4) --- how well the model preserves Yang Chaoyue's face when placed in a non-standard context.
  \item \texttt{historical\_authenticity}\,(0.4) --- accuracy of the 1970s attire and farming environment.
  \item \texttt{atmospheric\_tone}\,(0.2) --- consistency of the film grain and earth-tone color palette.
\end{itemize}
\noindent Every prompt additionally receives generic dimensions (prompt faithfulness, image quality, AI naturalness, composition). \\[2pt]
\textbf{Knowledge Gap References:} Textual knowledge of the background of 1970s Chinese history. Two reference images (period labor clothing; celebrity likeness) supplied alongside the generated image. \\[2pt]
{\footnotesize End-to-end trace and per-stage reasoner I/O: Appendix~\ref{app:reasoner_io}.}
\end{evalcontextbox}

\begin{table}[t]
\centering
\caption{\small\textbf{Evaluation components of the \bench{} judge.} Each component is scored 0--100; the overall score is the mean of all applicable components. Knowledge-sensitive components (top) vary per prompt and measure knowledge presence; rendering-quality components (bottom) are fixed across all prompts and measure generation competence independently of knowledge.}
\label{tab:eval_components}
\footnotesize
\begin{tabularx}{\columnwidth}{@{}lX@{}}
\toprule
\textbf{Component} & \textbf{What it evaluates} \\
\midrule
\multicolumn{2}{@{}l}{\emph{Knowledge-sensitive (prompt-adaptive)}} \\
Checklist Verification & 3--10 per-prompt checks on key visual elements, each scored on visual evidence \\
Rubric Scoring & 3--5 weighted scoring scheme that grades  quality across a few prompt-adaptive dimensions \\
Prompt Faithfulness & Presence and accuracy of all requested subjects, attributes, actions, scene, and style \\
Visual Reference Fidelity & Identity, attribute, and style fidelity to the prompt-requested entities (if applicable)\\
Textual Knowledge Fidelity & Correctness of factual/textual knowledge gaps (facts, systems, concepts, workflows,  etc) \\
\midrule
\multicolumn{2}{@{}l}{\emph{Rendering quality (knowledge-invariant)}} \\
Image Quality & Clarity and sharpness; freedom from artifacts; style, lighting, and perspective coherence \\
Text Rendering & Accuracy, readability, spelling, and placement of in-image text; scored only when text is required \\
AI Naturalness & Texture realism vs.\ AI smoothness and uncanny uniformity; grounded vs.\ generic look \\
Composition \& Aesthetics & Framing and balance; depth and spatial arrangement; color harmony; overall appeal \\
Physical Plausibility & Anatomy, object physics, spatial consistency, and material/lighting coherence \\
\bottomrule
\end{tabularx}
\vspace{-.3cm}
\end{table}

\subsection{What Makes World-Knowledge-Grounded Image Generation Hard?} 
\label{sec:bottleneck_finding}

\begin{finding}
\textbf{Finding 1: Generators that score comparably on standard prompts diverge by nearly 40 points when search-intensive world knowledge is required.}
On prompts requiring only parametric knowledge, open and commercial generators score at comparable levels (67--75 out of 100); on search-intensive prompts, the same open generators collapse to 22--28 while commercial systems with integrated search barely drop. This 40-point divergence is invisible to existing benchmarks that test only rendering within known concepts.
\end{finding}

\begin{figure}[t]
\centering
\includegraphics[width=\textwidth]{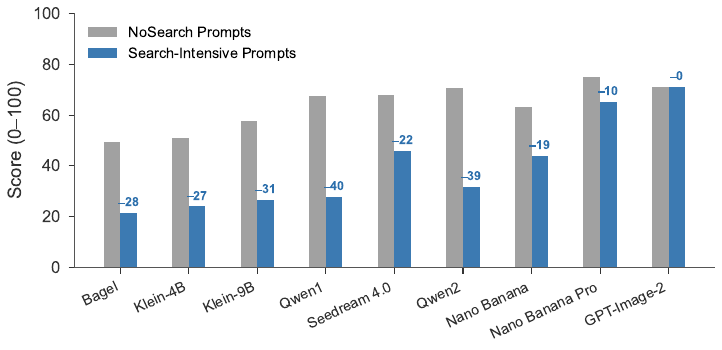}
\caption{\small\textbf{The world-knowledge bottleneck: per-stratum collapse across nine generators.}
Grouped bars show the nine-component mean on NoSearch vs.\ Search-Intensive strata.
Every generator, open-weight and commercial alike, collapses on the search-intensive subset,
confirming the bottleneck is universal. Drop magnitudes range from $-$0.1 (GPT-Image-2) to $-$39.1 (Qwen-Image-2).}
\label{fig:stratum_collapse}
\vspace{-.3cm}
\end{figure}

Figure~\ref{fig:stratum_collapse} visualizes the gap directly. On the NoSearch stratum, where only parametric knowledge is required, every generator, open and commercial alike, scores in a comparable band (63--75 out of 100). Generators cluster in a comparable band when knowledge is not the bottleneck. On the Search-Intensive stratum, the landscape fractures: every open-weight generator collapses to 22--28, while commercial systems with integrated search barely drop.

To isolate the world-knowledge bottleneck from rendering ability, we curate two complementary test sets: \emph{NoSearch} (100 prompts selected where generators perform well with parametric knowledge alone) and \emph{Search-Intensive} (651 prompts where external world knowledge is required).
Table~\ref{tab:bottleneck_overall} reports six key evaluation dimensions on the Search-Intensive slice, grouped into knowledge-sensitive components (which measure knowledge presence) and knowledge-invariant components (which measure rendering competence). The full nine-component breakdown including both strata appears in the appendix.

\begin{table}[t]
\centering
\caption{\textbf{The knowledge bottleneck on \bench{}: Search-Intensive subset (651 prompts).}
Scores on a 0--100 scale; higher is better.
\emph{Knowledge-sensitive} components (Checklist, Rubric, Visual Reference) vary per prompt and directly measure knowledge presence; \emph{knowledge-invariant} components (Text Rendering, Physical Plausibility, Image Quality) measure rendering competence.
Knowledge-sensitive components collapse while rendering components hold steady, confirming the failure is knowledge absence, not rendering inability.}
\label{tab:bottleneck_overall}
\small
\setlength{\tabcolsep}{4pt}
\rowcolors{2}{white}{rowgray}
\begin{adjustbox}{max width=0.92\columnwidth,center}
\begin{tabular}{@{}lccc|ccc@{}}
\toprule
& \multicolumn{3}{c}{\textit{Knowledge-Sensitive}} & \multicolumn{3}{c}{\textit{Knowledge-Invariant}} \\
\cmidrule(lr){2-4} \cmidrule(lr){5-7}
\textbf{Generator}
  & \textbf{Checklist} & \textbf{Rubric} & \textbf{Visual Reference}
  & \textbf{Text} & \textbf{Physical} & \textbf{Img Quality} \\
\midrule
\rowcolor{grouptop} \multicolumn{7}{@{}l}{\textit{\textbf{Open-weight generators}}} \\
Bagel~\citep{deng2025bagel}              & 18.2 & 17.6 & 13.5 & 2.5 & 36.8 & 30.5 \\
Flux.2-Klein-4B~\citep{blackforest2026flux2}    & 19.8 & 18.4 & 11.9 & 4.2 & 46.2 & 37.2 \\
Flux.2-Klein-9B~\citep{blackforest2026flux2}    & 24.2 & 23.1 & 16.9 & 7.2 & 48.6 & 36.8 \\
Qwen-Image~\citep{qwenimage}         & 24.8 & 24.3 & 17.7 & 8.7 & 44.6 & 40.1 \\
\midrule
\rowcolor{grouptop} \multicolumn{7}{@{}l}{\textit{\textbf{Commercial systems}}} \\
Qwen-Image-2~\citep{qwenimage}       & 28.5 & 27.1 & 21.0 & 12.7 & 48.2 & 42.2 \\
SeedDream-4.0~\citep{seedream2025arxiv}      & 44.2 & 43.6 & 35.1 & 35.9 & 64.0 & 57.0 \\
Nano Banana~\citep{nanobanana2025gemini}        & 41.0 & 40.4 & 33.2 & 28.0 & 65.5 & 57.1 \\
Nano Banana Pro~\citep{nanobanana2025gemini}    & 64.4 & 63.1 & 58.3 & 65.0 & 78.5 & 71.4 \\
GPT-Image-2~\citep{gptimage2openai}        & 71.2 & 70.1 & 66.0 & 75.9 & 77.3 & 75.1 \\
\bottomrule
\end{tabular}
\end{adjustbox}
\end{table}

The non-uniformity across commercial systems is itself diagnostic. GPT-Image-2 drops only 0.1 points, because live search operates behind that API. Nano Banana Pro drops only $9.7$ points ($75.0 \to 65.3$), while Qwen-Image-1 drops nearly $40$ points. The magnitude of the drop correlates inversely with the degree of integrated search capability: systems with reasoned search maintain performance, e.g., GPT-Image-2, Nano Banana Pro; systems without it collapse, e.g., open-source generators. 

Table~\ref{tab:bottleneck_overall} pinpoints where the gap originates. Knowledge-sensitive components (Checklist, Rubric, Visual Reference Fidelity), which directly measure knowledge presence, collapse across all open generators (e.g., Flux.2-Klein-9B: Checklist 24.2, Visual Reference 16.9). Knowledge-invariant components (Image Quality, Physical Plausibility) remain comparatively high (36.8, 48.6 for the same model), confirming that the collapse reflects knowledge absence, not rendering failure. GPT-Image-2 holds strong across both groups (Checklist 71.2, Visual Reference 66.0), consistent with integrated knowledge access.
The benchmark spans diverse, long-tailed domains (Figure~\ref{fig:bench_overview}), and the per-cell severity heatmap (Figure~\ref{fig:heatmap_severity}, Appendix~\ref{app:benchmark}) confirms the severity is pervasive across all twelve failure categories and twenty-two domains, ruling out the hypothesis that the bottleneck is concentrated in a few niche categories.

\subsection{Naive Search Is a Second Problem} 
\label{sec:naive}

The bottleneck has a natural remedy: search.
A human illustrator who receives an unfamiliar request looks up references before drawing.
The analogous pipeline for a generator has three steps: (1)~a reasoner identifies knowledge gaps in the prompt, (2)~search queries fill those gaps, returning images, text, or both, and (3)~the retrieved material is integrated into the generator's input so it can render faithfully (Figure~\ref{fig:pipeline}; \S\ref{sec:method} formalizes each step).

We stress-test this pipeline with two search policies.
\emph{BlindSearch}: the reasoner searches for every nominated gap regardless of whether search actually helps.
\emph{ReasonedSearch}: a frontier VLM (Gemini-3-Flash) selectively searches for quality-relevant gaps, then picks the better of the augmented and unaugmented outputs per prompt.

To diagnose when and where each policy helps or harms, we further partition the \emph{Search-Intensive} set into two subsets: \emph{VisualSearch} prompt set that benefits substantially from visual references, e.g., celebrities, or \emph{TextualSearch} that benefits from textual references, e.g., scientific infographics.


\begin{finding}
\textbf{Finding 2: Naive search actively degrades prompts the generator already handles.}
BlindSearch drops every generator on the NoSearch stratum (prompts that don't need external world knowledge); even selective search must learn \emph{when} to abstain (Table~\ref{tab:naive}).
\end{finding}

If search were uniformly beneficial, the problem would reduce to engineering better queries.
Table~\ref{tab:naive} shows otherwise.
On the \emph{NoSearch} stratum, \emph{BlindSearch} strictly degrades every generator: Qwen-Image-2 drops from $70.7$ to $60.4$, a $14.6\%$ relative loss on prompts it already handles well.
\emph{ReasonedSearch} improves Qwen-Image-2 on \emph{NoSearch} ($70.7 \to 76.5$) while strongly lifting \emph{TextualSearch} from $22.9$ to $34.1$.
Selective search buys knowledge-intensive gains while preserving parametric knowledge on prompts that do not benefit from retrieval.

\begin{table}[t]
    \centering
    \noindent
    \begin{minipage}[t]{0.475\linewidth}
      \vspace{0pt}%
      \centering
      \setlength{\tabcolsep}{2.5pt}
      \resizebox{\linewidth}{!}{%
      \small
      \begin{tabular}{@{}llccc@{}}
      \toprule
      \textbf{Need Search?} & \textbf{Generator}
        & \textbf{Baseline}
        & \textbf{Reasoned}
        & \textbf{Blind} \\
      \midrule
      \multirow{3}{*}{\footnotesize NoSearch}
      & Qwen-Image-2              & 70.7 & 76.5 & 60.4 \\
      & Qwen-Image             & 67.4 & 75.0 & 59.5 \\
      & Flux.2-Klein-9B & 57.8 & 66.4 & 52.3 \\
      \midrule
      \multirow{3}{*}{\footnotesize VisualSearch}
      & Qwen-Image-2              & 37.2 & 49.1 & 45.3 \\
      & Qwen-Image             & 32.8 & 44.4 & 39.7 \\
      & Flux.2-Klein-9B & 29.6 & 39.4 & 36.0 \\
      \midrule
      \multirow{3}{*}{\footnotesize TextualSearch}
      & Qwen-Image-2              & 22.9 & 34.1 & 32.1 \\
      & Qwen-Image              & 20.5 & 28.5 & 25.3 \\
      & Flux.2-Klein-9B & 18.9 & 26.6 & 24.0 \\
      \bottomrule
      \end{tabular}%
      }
      \captionof{table}{\small \textbf{Search is a double-edged sword.}
      We analyze three search polices on three knowledge  settings.}
      \label{tab:naive}
    \end{minipage}%
    \hfill
    \begin{minipage}[t]{0.47\linewidth}
      \vspace{0pt}%
      \centering
      \includegraphics[width=\linewidth]{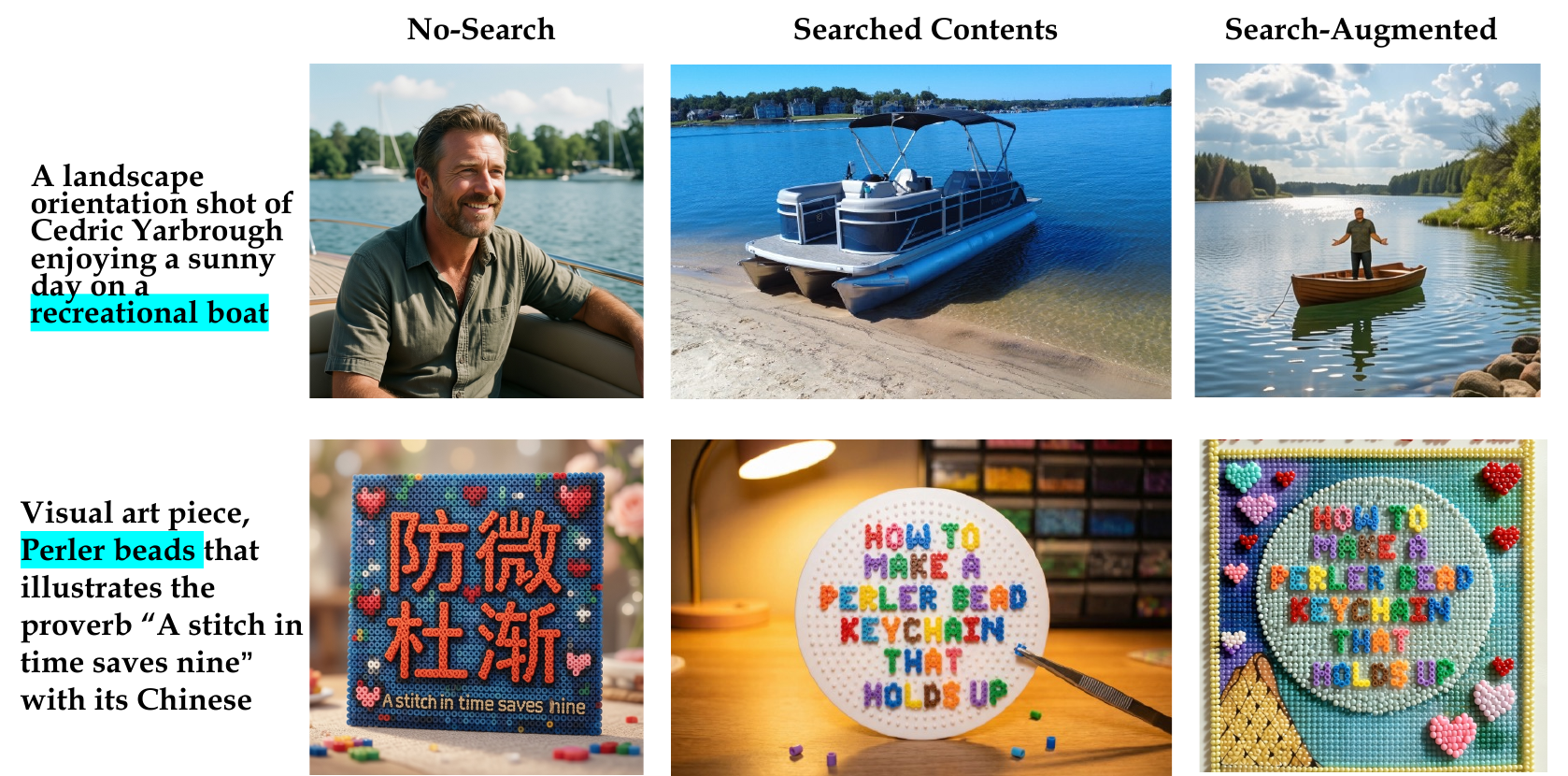}
      \captionof{figure}{\small \textbf{Failure examples of search.}
      Search queries executed are colored. Searched contents introduce noise and may degrade generation quality.}
      \label{fig:naive_failure}
    \end{minipage}
    \vspace{-.5cm}
  \end{table}

Figure~\ref{fig:naive_failure} illustrates two structurally distinct failure modes.
\emph{Concept corruption} (top): search fires on a prompt the generator already handles; the returned reference overrides accurate internal knowledge, a gating failure.
\emph{Copy effect} (bottom): a reference carries too much raw information and becomes a copying template rather than a knowledge supplement, a filtering failure.
These two modes directly motivate the gate--filter--integrate protocol in \S\ref{sec:atomic}.
The generator's inability to filter search noise mirrors text-domain evidence that parametric memory and external lookup trade off by query type~\cite{mallen2023nottrust}, but the visual setting adds modality-specific failure modes, including spatial distortion, identity blending, and stylistic leakage, that demand different triage and integration strategies.

\section{Co-Training {Agentic}  Visual Generation}
\label{sec:method}

The findings above motivate a search-augmented generation paradigm: search only when missing knowledge lies outside the generator's boundary, and control how retrieved content enters generation.
This requires deciding \emph{when} to search, \emph{which modality} to use, and \emph{how} to integrate evidence without introducing search noise.


We realize this philosophy through two components (Figure~\ref{fig:pipeline}): a reasoner controls what to search and how search-augmented inputs (visual references and searched web knowledge) reach the generator with noise resistance (\S\ref{sec:atomic}), and a co-training framework that first teaches the generator to expand knowledge, then recalibrates the reasoner to the strengthened generator's knowledge boundary (\S\ref{sec:training}).  

\subsection{Noise-Resistant {Agentic} Reasoner}
\label{sec:atomic}

Naive search degrades generation (\S\ref{sec:naive}) because noise enters at three structurally distinct points: the decision to search, the choice of reference, and the integration of external content.
A single filter cannot address all three; each demands its own targeted defense.
Therefore, we propose a three-stage agentic search protocol (bottom half of Figure~\ref{fig:pipeline}) that suppresses noise at each point in sequence (see Appendix~\ref{app:reasoner_io} for a worked example). {These stages form the agent's reason-and-act loop: \emph{gate} decides when to act and which tool to call, \emph{filter} selects the tool's returns, and \emph{integrate} assimilates the observation into generation.}

\textbf{Stage 1: Gate.}
In this stage, given a user prompt $p$, the reasoner identifies knowledge gaps, classifies each by type and severity, and proposes search queries with modality labels (\texttt{image}\,$|$\,\texttt{web}).
Only gaps rated \texttt{critical} or \texttt{important} trigger search; the rest are filtered, producing at most 3 search queries or \textsc{skip} if no actionable gaps survive.

\textbf{Stage 2: Filter.}
After multimodal search is executed and search results return, this stage aims to select references that sufficiently fill the specific gap identified  while minimizing extraneous content. We seek a reference that ``fills the gap'' supplies only the missing visual knowledge while leaving the generator free to compose the rest. Empirically, we find that this stage helps reduce the copy effects (\S\ref{sec:naive}) as visual shortcuts are filtered from search results.

\textbf{Stage 3: Integrate.}
Even a well-chosen visual reference can carry far more information than knowledge gap requires, and thus be exploited by the generator as visual shortcuts: raw pixel conditioning leaks style, background, and layout that were never requested. The integration stage therefore \emph{routes visual references through natural language} (Figure~\ref{fig:pipeline}): the reasoner fuses the original prompt, its gap analysis, retrieved visual and textual knowledge into enriched text specification. In addition to the raw photograph, the generator receives grounded citations naming exactly what to borrow, e.g., ``following Image~I, render the character in a teal-and-gold robe''. This preserves the missing knowledge for faithful generation while discarding the pixel-level noise that would otherwise dominate the output.

\subsection{Knowledge Boundary}
To address world-knowledge-grounded requests, our key insight is to introduce a reasoner that decides when and what to search, preparing augmented inputs to the generator. Yet this design rests on an implicit assumption: that there exists a well-defined partition between what the generator already knows and what it must search. The observations from \S\ref{sec:benchmark} confirm this partition is real but also reveal that it is \emph{generator-specific and shifts under training}; search helps some prompts and harms others, and the balance changes with generator capability. This identifies a quantity we define as \emph{knowledge boundary}.

\refstepcounter{definitionctr}
\begin{defbox} 
\textbf{Definition~\thedefinitionctr} (Knowledge Boundary).
\label{def:boundary}
Let $\mathcal{K}$ denote the space of world-knowledge units (entities, cultural symbols, typographic identities, procedural specifications, etc.)\ required by prompts in a distribution $\mathcal{P}$.
For a generator $G_\theta$ with parameters $\theta$, a prompt $p \in \mathcal{P}$, and conditioning context $c$ (search-returned references and text), let $Q(G_\theta, p, c) \in [0,1]$ be a bounded quality function.
For a fixed tolerance $\epsilon > 0$, define the \emph{internalizable} and \emph{contextual} knowledge sets:
\begin{align}
\mathcal{K}_{\mathrm{int}}(\theta) &\;=\; \bigl\{\, k \in \mathcal{K} \;:\; \mathbb{E}_{p \mid k}\bigl[Q(G_\theta,\, p,\, \text{\textsc{Search}}(k)) - Q(G_\theta,\, p,\, \varnothing)\bigr] < \epsilon \,\bigr\}, \label{eq:kint} \\
\mathcal{K}_{\mathrm{ctx}}(\theta) &\;=\; \mathcal{K} \;\setminus\; \mathcal{K}_{\mathrm{int}}(\theta). \label{eq:kctx}
\end{align}
where the expectation is over all prompts in $\mathcal{P}$ that require knowledge unit $k$, and $\text{\textsc{Search}}(k)$ denotes the conditioning context from executing the search protocol for knowledge unit $k$.
The pair $(\mathcal{K}_{\mathrm{int}}(\theta),\, \mathcal{K}_{\mathrm{ctx}}(\theta))$ forms a generator-specific partition of $\mathcal{K}$; we refer to this partition as the \emph{knowledge boundary} $\mathcal{B}(\theta)$.
A prompt $p$ is \emph{search-intensive} w.r.t.\ $G_\theta$ if it requires knowledge $k \in \mathcal{K}_{\mathrm{ctx}}(\theta)$.
The boundary is generator-specific: it depends on $\theta$ and shifts under training, i.e., $\mathcal{K}_{\mathrm{int}}(\theta) \subseteq \mathcal{K}_{\mathrm{int}}(\theta')$ when $\theta'$ results from DPO on search-augmented demonstrations.
\end{defbox}

Definition~\ref{def:boundary} formalizes our fundamental insight.
\begin{insight}
\textbf{Insight.} Some knowledge is internalizable and search should not fire for it; other knowledge is contextual and search is structurally necessary. 
\end{insight}
Crucially, the boundary need not be knowable a priori: we find that it is \emph{discoverable} through co-training. Figure~\ref{fig:cotraining_progression}b makes this concrete: we measure the per-prompt quality gap with and without search and show that its distribution shifts under co-training.

\begin{figure}[t]
\centering
\includegraphics[width=0.9\linewidth]{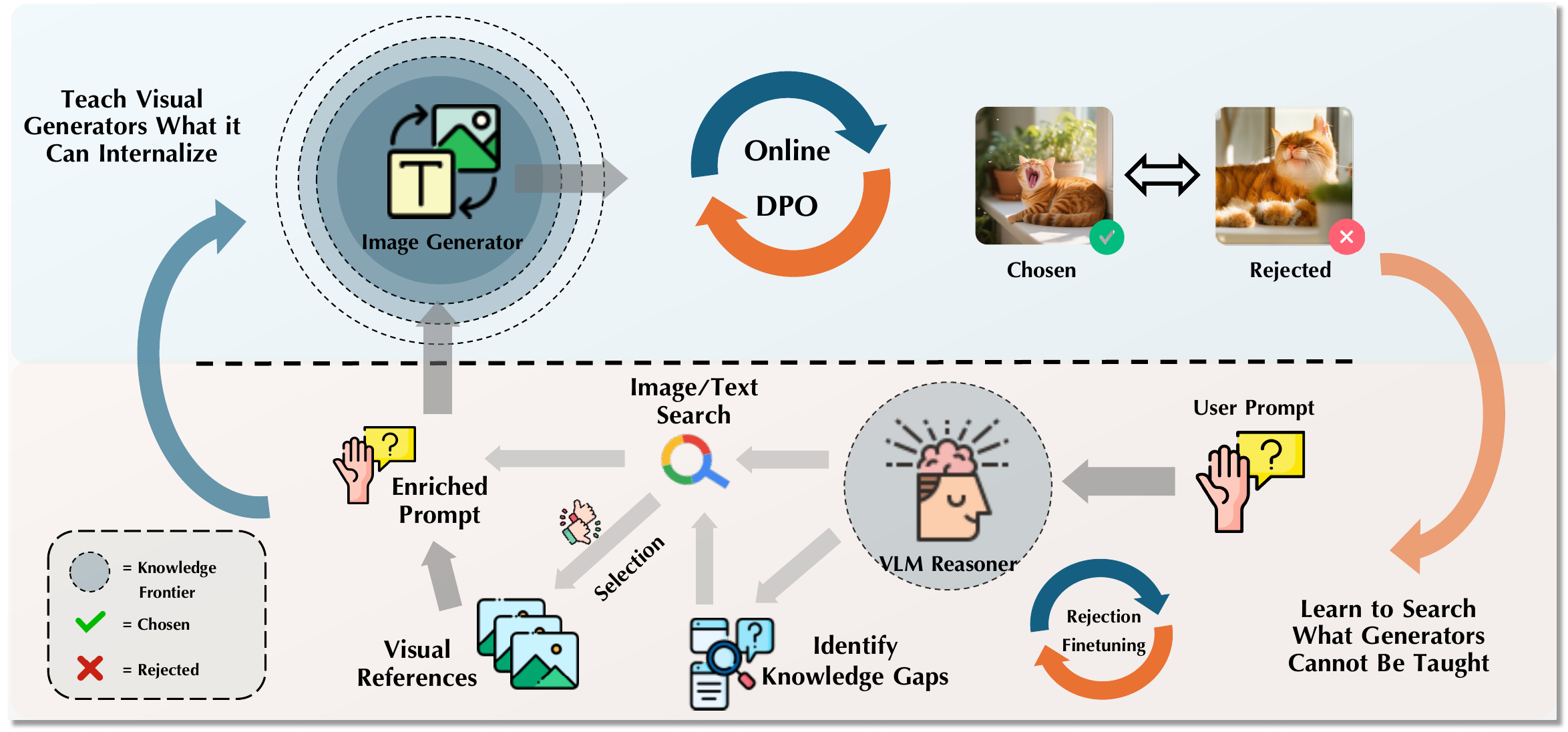}
\caption{\small\textbf{Co-training framework: teach the generator what it can internalize, then calibrate the reasoner to search what it cannot. }Given a user prompt, the VLM reasoner identifies knowledge gaps, executes modality-aware search (image or text), filters and integrates results into an enriched final prompt, and routes visual references to the generator. Co-training proceeds in two phases. Phase 1 (top): online DPO samples search-augmented generations and ranks them by generation quality, constructing preference pairs to push its knowledge boundary outward. Phase 2 (bottom): rejection-sampling finetuning recalibrates the reasoner to the strengthened generator, reinforcing trajectories where reasoned search improves output and discarding what causes degradation. }\vspace{-.3cm}
\label{fig:pipeline}
\end{figure}

\subsection{Co-Training: Teach, Then Search}
\label{sec:training}

How does co-training discover the boundary in practice? The boundary shifts only when the generator's parameters change, so discovering it requires training the generator itself, not merely adjusting the search policy.
Co-training (top half of Figure~\ref{fig:pipeline}) acts on both sides in sequence: it first strengthens the generator so fewer prompts require search, then recalibrates the reasoner to the strengthened generator's boundary.


\textbf{Phase~0: Supervised warm-start.}
Co-training requires a reasoner that follow the prescribed three-stage protocol to reason and search.
So we first warm-up the reasoner by supervised finetuning Qwen3-VL-8B~\cite{qwen2024vl} on ${\sim}$10{,}000 expert-annotated Task~A/B/C trajectories, which correspond one-to-one to the gate/filter/integrate stages (representative per-stage I/O in Appendix~\ref{app:reasoner_io}; full hyperparameters in Appendix~\ref{app:hardware}).
The resulting reasoner produces well-structured analyses but remains generator-agnostic: it searches for anything that \emph{might} be difficult, not only what \emph{one specific} generator actually fails on.

\begin{figure}[t]
\centering
\includegraphics[width=\textwidth]{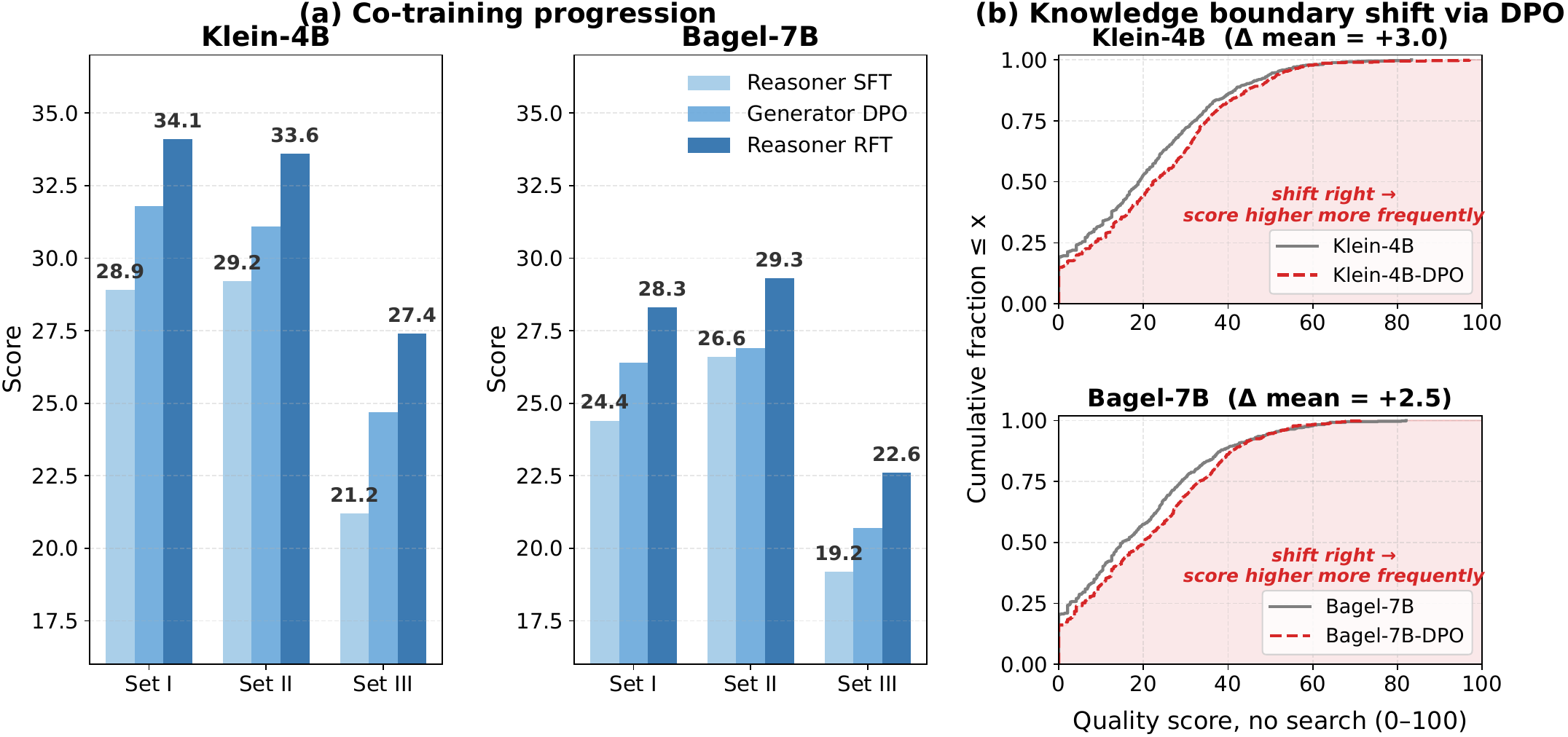}
\caption{\small\textbf{Co-training progression and knowledge boundary shift.}
\textbf{(a)} Grouped bars show the three co-training stages (Reasoner SFT, Generator DPO, Reasoner RFT) for Set~I (easiest), Set~II, and Set~III (hardest) search-intensive tiers (Klein-4B). All three tiers show monotonic improvement.
\textbf{(b)} Cumulative distribution of per-prompt no-search quality scores for base vs.\ DPO-finetuned generators on the 647-prompt eval set. The DPO curve sits below the base curve, indicating a rightward shift: fewer prompts receive low scores and more receive high scores from parametric knowledge alone. The shaded region represents newly internalized knowledge: concepts that previously required search but now reside in generator parameters. The shift is consistent across both Klein-4B (top) and Bagel-7B (bottom), confirming that DPO expands the knowledge boundary regardless of architecture.}
\label{fig:cotraining_progression}
\vspace{-.3cm}
\end{figure}

\textbf{Phase~1: Teach what the generator can internalize.}
The warmed-up reasoner provides search-augmented inputs (visual references and text prompts), exposing the generator to knowledge it previously lacked.
Based on original text prompts and augmented inputs, we train generators to internalize world knowledge via online iterative Diffusion-DPO~\cite{rafailov2023dpo,wallace2024diffusiondpo}. In each episode, the generator samples $M$ images per prompt, scored by the \dataset{} evaluation protocol, and constructs preference pairs from top- and worst-scored generations.
DPO reinforces the generator's own best outputs, internalizing knowledge whose visual identity is stable enough to absorb into parameters.
After Phase~1, the generator $G_\theta^{(1)}$ has expanded its knowledge boundary: concepts that previously required search now reside in parameters (Figure~\ref{fig:cotraining_progression}b provides direct evidence: the CDF of per-prompt no-search quality shifts rightward after DPO).

DPO training also builds noise-robustness in visual generation: because the generator trains on search-augmented inputs, it learns to incorporate imperfect visual references without being dominated by their noise. A generator trained only on clean inputs treats every conditioning signal as authoritative, the root cause of concept corruption and copy effects identified in \S\ref{sec:naive}. Exposure to the noisy, varied outputs of real search builds robustness to imperfect references. This noise resistance is essential for Phase~2, because even a well-calibrated reasoner cannot guarantee perfectly clean references at inference time.

\textbf{Phase~2: Search what the generator cannot be taught.}
The generator's boundary has shifted; the reasoner's policy is now miscalibrated, still searching for concepts the generator has internalized.
Recalibration uses rejection-sampling finetuning (RFT): we roll out $N$ trajectories from the Phase~0 reasoner paired with $G_\theta^{(1)}$, score the resulting images, compute group-relative advantages, and retain only positive-advantage trajectories for continued reasoner training.
Trajectories where the reasoner correctly abstains receive high scores and are reinforced; trajectories where unnecessary search corrupts output receive low scores and are discarded.
The reasoner learns the strengthened generator's boundary without explicit boundary labels: the boundary emerges from the reward signal. {We roll out the search agent's trace, score the resulting images, and reinforce positive-advantage trajectories, optimizing the agent's policy against the generator as its environment.}
The full RFT cycle fits in $4 \times 8$ GPU hours.

Taken together, the two phases execute complementary movements on the knowledge boundary: Phase~1 moves the boundary outward by internalizing stable knowledge and building noise robustness; Phase~2 moves the search policy inward to match, so that search fires precisely where knowledge remains outside the generator's expanded capacity.


\begin{algorithm}[h]
\caption{Co-Training: Teach, Then Search}
\label{alg:cotrain}
\begin{algorithmic}[1]
\Require Generator $G_\theta$, base VLM $R_\phi$, dataset $\mathcal{D} = \{p_i\}_{i=1}^{N_{\mathrm{D}}}$, quality function $Q$, margin $\epsilon$
\Ensure Co-trained generator $G_{\theta'}$, calibrated reasoner $R_{\phi'}$

\Statex \textbf{--- Phase~0: Supervised Warm-Start ---}
\State Collect expert trajectories $\mathcal{T}_{\mathrm{sft}} = \{(p_i, a_i^{\mathrm{A}}, a_i^{\mathrm{B}}, a_i^{\mathrm{C}})\}$ for Tasks A/B/C \Comment{${\sim}$20K}
\State $R_{\phi_0} \leftarrow \arg\min_\phi \sum_{(p,a) \in \mathcal{T}_{\mathrm{sft}}} \mathcal{L}_{\mathrm{SFT}}(\phi; p, a)$
    \Comment{Cross-entropy on structured task outputs}
\Statex

\Statex \textbf{--- Phase~1: Online Iterative DPO (Generator) ---}
\For{each episode $e = 1, \dots, T_1$}
    \State Sample batch $\mathcal{B}_e \subset \mathcal{D}$
    \For{each $p \in \mathcal{B}_e$}
        \State $\tilde{p} \leftarrow R_{\phi_0}(p)$ \Comment{Search-enriched specification via gate--filter--integrate}
        \State Generate $M$ images: $\{x_j\}_{j=1}^{M} \sim G_{\theta}(\cdot \mid \tilde{p})$
        \State Score: $s_j \leftarrow \text{Score}(p, x_j)$ for each $j$ \Comment{Image scoring function; distinct from boundary-defining $Q$ in Def.~\ref{def:boundary}}
        \State Construct preference pair: $x^{w} \leftarrow x_{\arg\max s_j}$,\; $x^{l} \leftarrow x_{\arg\min s_j}$
    \EndFor
    \State Update via DPO loss adapted for flow-matching:
    \Statex \qquad $\mathcal{L}_{\mathrm{DPO}}(\theta) = -\mathbb{E}\!\left[\log\sigma\!\left(\beta \left(\log \frac{v_\theta(x^w_t \mid \tilde{p})}{v_{\theta_{\mathrm{ref}}}(x^w_t \mid \tilde{p})} - \log \frac{v_\theta(x^l_t \mid \tilde{p})}{v_{\theta_{\mathrm{ref}}}(x^l_t \mid \tilde{p})}\right)\right)\right]$
    \Statex \qquad where $v_\theta$ is the flow-matching velocity field, $t$ is the diffusion timestep, $\beta = 100$ is the DPO temperature, and $\theta_{\mathrm{ref}}$ is updated via EMA (decay $0.99$). Preference pairs are constructed from groups of $M$ generations per prompt.
    \State $\theta \leftarrow \theta - \eta_1 \nabla_\theta \mathcal{L}_{\mathrm{DPO}}(\theta)$
\EndFor
\State $G_{\theta'} \leftarrow G_\theta$ \Comment{DPO-strengthened generator}
\Statex

\Statex \textbf{--- Phase~2: Rejection Finetuning (Reasoner) ---}
\For{each $p \in \mathcal{D}_{\mathrm{rft}} \subset \mathcal{D}$}
    \State Roll out $N_{\mathrm{traj}}$ trajectories: $\{\tau_n\}_{n=1}^{N_{\mathrm{traj}}}$, where $\tau_n = R_{\phi_0}(p)$, each yielding image $x_n \sim G_{\theta'}(\cdot \mid \tau_n)$
    \State Score: $s_n \leftarrow \text{Score}(p, x_n)$
    \State Compute group-relative advantage: $A_n \leftarrow \frac{s_n - \bar{s}}{\sigma_s + \delta}$, \; $\bar{s} = \frac{1}{N_{\mathrm{traj}}}\sum_n s_n$, \; $\sigma_s = \mathrm{std}(\{s_n\})$
\EndFor
\State Filter: $\mathcal{T}_{\mathrm{rft}} \leftarrow \{\tau_n : A_n > 0\}$ \Comment{Retain positive-advantage trajectories}
\State $R_{\phi'} \leftarrow \arg\min_\phi \sum_{\tau \in \mathcal{T}_{\mathrm{rft}}} \mathcal{L}_{\mathrm{SFT}}(\phi; \tau)$
    \Comment{SFT on filtered trajectories}
\Statex
\State \Return $(G_{\theta'},\; R_{\phi'})$
\end{algorithmic}
\end{algorithm}


\section{Validating Evolving Knowledge Boundary through Co-Training}
\label{sec:experiments}

If the knowledge boundary is real and co-training discovers it, three consequences follow.
First, each training phase should improve overall quality without regressing on any stratum; \emph{monotonicity} rules out the possibility that gains on search-intensive prompts come at the cost of prompts the generator already handles.
Second, the improvement should be \emph{selective}: the calibrated reasoner must recover performance on prompts that generic search degrades, demonstrating that it has learned \emph{when not to search}.
Third, the optimal search policy should be \emph{generator-specific}: a policy calibrated to the base generator should behave differently from one calibrated to the DPO-strengthened generator, confirming that the boundary is a joint property of the generator--reasoner pair rather than a fixed property of the prompt distribution.
We test each prediction below; per-dimension breakdowns and the broader generator landscape appear in the appendix.


\textbf{Generators.}
Co-training experiments use two architecturally distinct open-weight generators: Flux.2-Klein-4B (\textbf{Klein-4B}), a flow-matching model, and Bagel-7B (\textbf{Bagel}), a unified vision--language model. This model choice is deliberate because the two models are widely acknowledged open-source generators that support both text-to-image generation and reference-to-image generation. 
Testing on two architectures rules out the possibility that co-training benefits are specific to one conditioning mechanism.
Each has a DPO-finetuned variant (\textbf{Klein-4B-DPO} and \textbf{Bagel-DPO}) whose knowledge boundary has shifted outward through Phase~1 training.

\textbf{Reasoner variants.}
The progression from weaker to stronger reasoners isolates the effect of generator-aware calibration:
\textbf{\CondNS}: generator alone (no retrieval baseline).
\textbf{\CondGS} (SFT-8B): Qwen3-VL-8B after supervised warm-up; the reasoner follows the protocol to perform reasoning and search but is generator-agnostic, i.e., not aware of knowledge boundary of specific generators.
\textbf{\CondGAS}: undergoes co-training based on \CondGS; the reasoner is calibrated to specific generator's specific boundary.
\textbf{\textsc{Oracle} (Frontier API)}: trillion parameters high-cost Gemini-3-Flash API as the reasoner following same reasoning and search protocol. This serves as an upper bound on generator-agnostic reasoner, showing best possible performance when pairing generators with strong but generic reasoner.

All reasoners are equipped with search tools for world-knowledge grounding. We use Google Image and Web Search via SERP API. 

\textbf{Evaluation.}
All scores use the Gemini-3-Flash judge on the 751-prompt test set (0--100 scale, 9-component rubric averaged to an overall score; Appendix~\ref{app:eval}).
The search-intensive prompts are partitioned into three difficulty sets by Nano Banana Pro no-search quality (tercile boundaries): Set~I (easiest), Set~II, and Set~III (hardest).  

\vspace{-.1cm}
\subsection{Co-Training Produces Monotonic, Selective Improvement}
\label{sec:exp_monotone}
\label{sec:main_results}
\label{sec:exp_strata}

With the setup in place, we first test our key insight: that co-training \emph{discovers and expands} the generator's knowledge boundary. Figure~\ref{fig:cotraining_progression} supplies the two most direct pieces of visual evidence. Panel~(a) shows the co-training progression---Reasoner SFT, Generator DPO, Reasoner RFT---improving monotonically within every difficulty tier, previewing the downstream quality gains we quantify below. Panel~(b) exposes the underlying mechanism: the CDF of per-prompt no-search quality for the base vs.\ DPO-strengthened generator shifts rightward, meaning fewer prompts score low and more score high from parametric knowledge alone. The shaded region is newly internalized knowledge---concepts whose per-prompt quality gap formerly exceeded $\epsilon$ and now fall below it, migrating from $\mathcal{K}_{\mathrm{ctx}}$ to $\mathcal{K}_{\mathrm{int}}$ (Def.~\ref{def:boundary}). The shift holds across both Klein-4B and Bagel-7B, confirming that DPO expands the boundary regardless of architecture.

\begin{finding} 
\textbf{Finding 3: Co-training discovers the knowledge boundary: a calibrated 8B reasoner matches a frontier oracle on the same generator.}
The monotonic progression from \CondNS{} through \CondGS{} to \CondGAS{}, the per-stratum recovery on NoSearch prompts, and the generator-specific behavior across two architectures provide converging evidence that the structural split is an operational design variable, not merely a conceptual framework (Table~\ref{tab:main}).
\end{finding}

Table~\ref{tab:main} quantifies this progression.
The top section traces three phases in order: Phase~0 pairs a generator-agnostic SFT reasoner with the base generator; Phase~1 keeps the same reasoner but pairs it with the DPO-strengthened generator (isolating the generator improvement); Phase~2 recalibrates the reasoner to the strengthened generator via RFT (isolating the reasoner calibration effect).
The bottom section provides reference baselines that bound interpretation: no-search baselines establish the floor, frontier oracle baselines establish the ceiling, and a cross-check (calibrated reasoner on base generator) tests generator-specificity directly.

\begin{table}[t]
\centering
\caption{\small \textbf{Main results: the co-training progression.}
The \textbf{Progression} section traces the three co-training phases for both Klein-4B and Bagel.
The \textbf{Reference baselines} section reports the no-search floor, frontier oracle ceiling, and cross-generator check for each architecture.
Difficulty terciles (Set~I--III) by Nano Banana Pro no-search quality; NoSearch subset (100 prompts) reported separately.}
\label{tab:main}
\small
\resizebox{\columnwidth}{!}{%
\begin{tabular}{@{}llccccc@{}}
\toprule
& & NoSearch & \multicolumn{4}{c}{Search} \\
\cmidrule{4-7}
& & & Set I & Set II & Set III & Overall \\
\midrule
\multicolumn{7}{l}{\textbf{Co-training progression}} \\
\midrule
\multicolumn{7}{l}{\quad\textit{Klein-4B}} \\
Phase~0 & \CondGS\,(SFT-8B)\;+\;Klein-4B & 54.6 & 28.9 & 29.2 & 21.2 & 26.4 \\
Phase~1 & \CondGS\,(SFT-8B)\;+\;Klein-4B-DPO & 54.0 & 31.8 & 31.1 & 24.7 & 29.2 \\
Phase~2 & \CondGAS\,(RFT-8B)\;+\;Klein-4B-DPO & \textbf{56.9} & \textbf{34.1} & \textbf{33.6} & \textbf{27.4} & \textbf{31.8} \\
\addlinespace
\multicolumn{7}{l}{\quad\textit{Bagel}} \\
Phase~0 & \CondGS\,(SFT-8B)\;+\;Bagel & 52.6 & 24.4 & 26.6 & 19.2 & 23.4 \\
Phase~1 & \CondGS\,(SFT-8B)\;+\;Bagel-DPO & 52.4 & 26.4 & 26.9 & 20.7 & 24.7 \\
Phase~2 & \CondGAS\,(RFT-8B)\;+\;Bagel-DPO & \textbf{54.3} & \textbf{28.3} & \textbf{29.3} & \textbf{22.6} & \textbf{26.8} \\
\midrule
\multicolumn{7}{l}{\textbf{Reference baselines}} \\
\midrule
\multicolumn{7}{l}{\quad\textit{Klein-4B}} \\
& \CondNS\;+\;Klein-4B-DPO & 49.9 & 28.2 & 26.3 & 20.6 & 25.0 \\
& \textsc{Oracle}\;+\;Klein-4B-DPO & 55.7 & 33.7 & 33.9 & 26.0 & 31.2 \\
& \CondGAS\,(RFT-8B)\;+\;Klein-4B & 54.6 & 29.0 & 29.8 & 21.5 & 26.8 \\
\addlinespace
\multicolumn{7}{l}{\quad\textit{Bagel}} \\
& \CondNS\;+\;Bagel-DPO & 48.8 & 24.3 & 24.0 & 18.8 & 22.4 \\
& \textsc{Oracle}\;+\;Bagel-DPO & 53.2 & 28.1 & 29.0 & 21.1 & 26.1 \\
& \CondGAS\,(RFT-8B)\;+\;Bagel & 52.5 & 25.1 & 27.3 & 19.8 & 24.1 \\
\bottomrule
\end{tabular}%
}
\vspace{-.3cm}
\end{table}

Three patterns emerge, each confirming a prediction of the co-training principle.

\textbf{Monotonicity.}
The progression is monotonic on both generators. On Klein-4B, Phase~2 reaches 31.8, slightly exceeding the frontier oracle on the same generator (31.2). Both training phases contribute independently: generator DPO alone (Phase~0$\to$1) yields $+2.8$; reasoner RFT (Phase~1$\to$2) adds $+2.6$. The same pattern holds for Bagel (Table~\ref{tab:main}).

As Figure~\ref{fig:cotraining_progression}(a) confirms, the improvement is monotonic within every difficulty tier, with Set~III (hardest) showing the largest gains where the boundary shift (Figure~\ref{fig:cotraining_progression}b) has the most room to operate.

\textbf{Selectivity.}
Recall from \S\ref{sec:naive} that naive search \emph{degrades} prompts the generator already handles (Finding~2).
The calibrated reasoner must therefore improve search-intensive prompts without sacrificing quality on prompts where search is harmful.
The NoSearch column in Table~\ref{tab:main} tests this directly: Phase~2 scores 56.9, a $+$7.0 gain over the no-search DPO baseline (49.9).
The reasoner has learned when to abstain; it no longer corrupts prompts the strengthened generator already handles.
Within the Search partition, Phase~2 improves over the no-search DPO baseline at every set and matches or closely approaches the frontier oracle, with the largest gains on Set~III where external knowledge is most critical.
The same selectivity pattern holds for Bagel (Phase~2: 54.3 on NoSearch vs.\ 48.8 for no-search DPO; 26.8 overall).

\textbf{Generator-specificity.}
If the knowledge boundary is generator-specific, a reasoner calibrated to the DPO-strengthened generator should behave differently when paired with the base generator, because the base generator's boundary sits in a different location.
The cross-check rows in Table~\ref{tab:main} test this: \CondGAS{} paired with \emph{base} Klein-4B scores 26.8 overall (vs.\ 31.8 when paired with Klein-4B-DPO, the generator it was calibrated for).
NoSearch recovery remains strong (the reasoner still knows when to abstain), but Set~III gains shrink because the reasoner searches for concepts the base generator cannot render even with correct references.
The calibrated reasoner, when paired with its \emph{intended} generator, matches or exceeds frontier-oracle scores at every tier.
This asymmetry confirms that the search policy is a joint property of the generator--reasoner pair: a policy optimal for one generator is suboptimal for another.

\textbf{Structural interpretation.}
The evidence confirms all three predictions: monotonicity across phases and generators, selectivity on the NoSearch stratum, and generator-specificity in the cross-check.
The matched-compute comparison is the sharpest validation: our co-trained 8B reasoner paired with a 4B generator (31.8) slightly exceeds a frontier VLM oracle on the same generator (31.2), suggesting that generator-specific calibration can approach costly frontier-scale reasoning at a fraction of the compute, latency and expense.
The 39-point gap to GPT-Image-2 (71.0) reflects generator capacity at the 4B scale rather than a framework limitation; the co-training principle extracts the maximum from a fixed generator, and scaling the generator along this axis is the clear path to closing the absolute gap.

\section{Related Work}
\label{sec:related}

\textbf{Search-augmented generation.}
Retrieval-augmented generation (RAG)~\cite{lewis2020rag} and REALM~\cite{guu2020realm} established the paradigm of grounding language models in external documents at inference time.
Atlas~\cite{izacard2023atlas} and REPLUG~\cite{shi2023replug} take this further by jointly training the searcher and language model end-to-end, demonstrating that aligning the retrieval distribution with the generator's needs yields substantial gains over frozen-searcher pipelines.
In the visual domain, Re-Imagen~\cite{chen2023reimagen} conditions diffusion on searched image--text pairs, RDM~\cite{blattmann2022rdm} augments generation with nearest-neighbor features, KNN-Diffusion~\cite{ashual2023knn} queries a large image database at generation time, Gen-Searcher~\cite{feng2026gen} trains an agentic searcher via SFT and RL while keeping the image generator frozen, and IP-Adapter~\cite{ye2023ipadapter} injects visual exemplars through cross-attention adapters.
Subject-driven methods such as DreamBooth~\cite{ruiz2023dreambooth} and Textual Inversion~\cite{gal2022textual} internalize specific concepts via per-concept optimization; BLIP-Diffusion~\cite{li2023blipdiffusion} pre-trains a subject representation for efficient control.
A common thread unites these pipelines: search is \emph{always on}, triggering retrieval for every prompt regardless of whether the generator already possesses adequate knowledge.
In contrast, we study real user requests from production AIGC systems, identify real-world failure mode taxonomy, and contribute datasets and replayable search harness with  this real-world diversity. Our analysis (\S\ref{sec:benchmark})  shows that indiscriminate search actively degrades a substantial fraction of prompts, making the \emph{decision to search} itself a first-order design variable. Our approach thus proposes to co-evolve the joint system of agentic reasoner and image generator, laying the foundation for recursive self-improvement in visual generation. 

\textbf{Selective and adaptive retrieval.}
The question of \emph{when} to search has been studied extensively in text.
Self-RAG~\cite{asai2024selfrag} trains a language model to emit retrieval-control and critique tokens.
FLARE~\cite{jiang2023flare} triggers retrieval when predictive confidence drops.
Toolformer~\cite{schick2023toolformer} learns when to invoke external APIs including search.
CRAG~\cite{yan2024crag} evaluates and corrects retrieval quality post-hoc.
Adaptive-RAG~\cite{jeong2024adaptiverag} routes queries to different retrieval strategies based on complexity.
Mallen et al.~\cite{mallen2023nottrust} and Kandpal et al.~\cite{kandpal2023entities} show that popular entities are better served by parametric knowledge while rare entities benefit from retrieval, an empirical split analogous to our internalizable/contextual distinction.
These methods establish selective retrieval for text, but visual search failure is structurally different: it manifests as spatial distortion, identity blending, and stylistic contamination (\S\ref{sec:naive}), failure modes that have no text-domain analogue and require modality-specific triage and integration.
More fundamentally, none co-trains the search policy with the generator: the retrieval decision is learned against a fixed generation model, whereas our co-training framework recalibrates the search policy after each generator improvement.

\textbf{{Tool-augmented and agentic} image generation.}
GenAgent~\cite{genagent2024}, RPG-DiffusionMaster~\cite{yang2024rpg}, LLM-grounded Diffusion~\cite{lian2024lmd}, and IterComp~\cite{zhang2024itercomp} use LLM-based planning to decompose complex prompts, improving compositional generation within the generator's existing knowledge.
Instruct-Imagen~\cite{hu2024instructimagen} supports multi-modal instructions including style exemplars. RationalRewards~\cite{wang2026rationalrewards} show that preference-calibrated prompt rewriter significantly improve image generation via agentic workflow at test-time.
These systems assume the generator possesses the relevant visual concepts and focus on compositional arrangement; none addresses knowledge \emph{absence}, the case where the generator has never seen the concept it must render.
Our work is complementary: agentic planning can improve the compositional quality of search-augmented outputs, but cannot substitute for the missing knowledge that search provides.

\textbf{Knowledge-intensive evaluation.}
Existing benchmarks evaluate compositionality and aesthetic quality within known visual concepts: GenAI-Bench~\cite{genaibench} tests prompt adherence, T2I-CompBench~\cite{t2icompbench} evaluates spatial reasoning and attribute binding, DALL-Eval~\cite{dalleval} measures object relationships, and HEIM~\cite{lee2024heim} provides holistic multi-dimensional evaluation.
VLM-based automated judges such as TIFA~\cite{hu2023tifa} and human-grounded auto-metric suites such as Gecko~\cite{wiles2024gecko} enable scalable faithfulness assessment.
A prompt like ``a red cube on a blue sphere'' tests spatial reasoning but assumes the generator knows what cubes and spheres look like.
\bench{} is complementary: it targets concepts \emph{outside} the training distribution (cultural symbols, recent entities, niche typography), measuring knowledge absence rather than reasoning failure.
The gap is stark: Flux.2-Klein-9B scores above 4.0/5.0 on GenAI-Bench yet falls below 30/100 on \bench{}, a discrepancy of over 60 percentage points that no existing evaluation captures.

\textbf{Self-improving and co-training systems.}
Self-Rewarding Language Models~\cite{yuan2024selfrewarding} show that a model can judge its own outputs to generate training signal, improving over successive iterations without external reward.
SPIN~\cite{chen2024spin} frames self-improvement as self-play, iteratively distinguishing model outputs from ground truth. Our co-training framework instantiates this principle for visual generation: Phase~1 (DPO) teaches the generator from search-augmented demonstrations that expose knowledge it previously lacked, while Phase~2 (RFT) recalibrates the reasoner to the strengthened generator's shifted boundary.
The deliberate ordering (teach, then search) and the boundary-discovery mechanism distinguish our approach from generic self-play, where both players improve symmetrically; here the generator and reasoner occupy structurally different roles and improve in complementary directions.

\section{Conclusion}
\label{sec:conclusion}

The central question is not how to build a model that knows everything, but how to build a system that knows what it does not know. This split creates a generator-specific knowledge boundary, and we have shown that this boundary, though hard to specify a priori, is discoverable through co-training. Our evidence suggests that this self-awareness emerges from co-training dynamics, not from scale alone.

The resolution lies in co-training the generator and reasoner around their shared boundary. Online DPO serves a dual function: it pushes the generator's boundary outward by internalizing stable visual knowledge, and it teaches the generator to use imperfect references without being corrupted by them. Rejection finetuning then recalibrates the reasoner to search only what the strengthened generator still cannot render. The monotonic progression from \CondNS{} through \CondGS{} to \CondGAS{}, the per-stratum recovery on NoSearch prompts, and the generator-specific behavior across two architectures provide converging evidence that the structural split is an operational design variable, not merely a conceptual framework.

Our deliberately minimal recipe (one DPO pass, one RFT pass, a 4B generator, an 8B reasoner) validates the principle at the smallest useful scale. Each subsequent iteration can push the boundary further outward and tighten the search policy further inward, converging toward a regime where only genuinely contextual knowledge triggers search. A natural objection is that next-generation models will absorb more knowledge at scale, making external search unnecessary. But training data is finite while the world is unbounded and in constant motion: no model, regardless of scale, can internalize knowledge about events after its training cutoff, entities too rare for any feasible dataset, or cultural knowledge that evolves across communities and time. The knowledge boundary may shift outward with scale, but it cannot disappear; co-training discovers where it lies for any given generator, at any scale. Whether additional iterations sharpen the boundary, whether larger generators internalize knowledge that resists parameterization in smaller ones, and whether richer reward signals accelerate gains on the hardest strata are open questions that define a clear research trajectory.

Beyond these immediate questions, this work opens several broader research directions. What is the scaling law of the knowledge boundary: does the internalizable set expand uniformly across failure categories as generator size increases, or do certain categories (e.g., cultural specificity) resist internalization? Can the boundary be predicted from model internals without full co-training, for instance via probing classifiers or uncertainty estimates? Does the internalizable/contextual split generalize to other modalities such as video, 3D, and music generation, which face the same structural problem of finite training data and unbounded user requests?

{\textbf{Beyond search: the knowledge boundary as a tool-invocation principle.} Search is the first and most fundamental tool, but the boundary that governs \emph{when to invoke it} is not search-specific. The same gate--filter--integrate protocol and teach-then-search co-training extend to other tools---image editing and inpainting, render-as-code (SVG, matplotlib, CAD), 3D-asset retrieval, and structural control---each supplying a different slice of the contextual knowledge set $\mathcal{K}_{\mathrm{ctx}}$. Whether a single agent can learn a unified boundary across such a toolbox is an open direction this work's harness is built to support.}

The released dataset, co-training corpus, and search corpus provide the infrastructure for pursuing these questions: 20{,}939 prompt records (20{,}789 unique prompt strings), 96{,}848 reasoning trajectories, 283{,}493 generated images, and 159{,}027 archived search sessions that can be replayed without live search API access.

\bibliographystyle{colm2024_conference}
\bibliography{references}

\clearpage
\appendix
\section*{Appendix overview}
{This appendix is organized to follow the evidence chain of the paper.}
{\textbf{\S\ref{app:benchmark}} documents benchmark construction and the per-category/domain heatmap used by the main text.}
{\textbf{\S\ref{app:failure_ext}--\S\ref{app:eval_correlation}} provide focused failure-analysis context and judge--human agreement.}
\textbf{\S\ref{app:full_component_breakdown}} gives the full nine-component score breakdown across all generators.
\textbf{\S\ref{app:prompts}--\S\ref{app:eval}} provide the end-to-end workflow, reasoner I/O examples, and evaluation protocol.
\textbf{\S\ref{app:hardware}} documents hardware and training/serving implementation details.
\textbf{\S\ref{app:limitations}--\S\ref{app:broader}} summarize limitations, release safeguards, third-party/API caveats, and broader impact.
\section{Benchmark and dataset details}
\subsection{Benchmark construction details}
\label{app:benchmark}
\textbf{Taxonomy annotation reliability.}
Two annotators independently labeled 2{,}000 sampled failure cases, iteratively
refined the schema to Cohen's $\kappa > 0.85$, then applied it to all
10{,}840 prompts in the production scrape.
The taxonomy's stability across annotators supports viewing it as capturing
structure in the failure space rather than an artifact of labeling choices.

The failure taxonomy appears in Table~\ref{tab:taxonomy} (main paper).

\begin{figure}[t]
\centering
\includegraphics[width=0.85\linewidth]{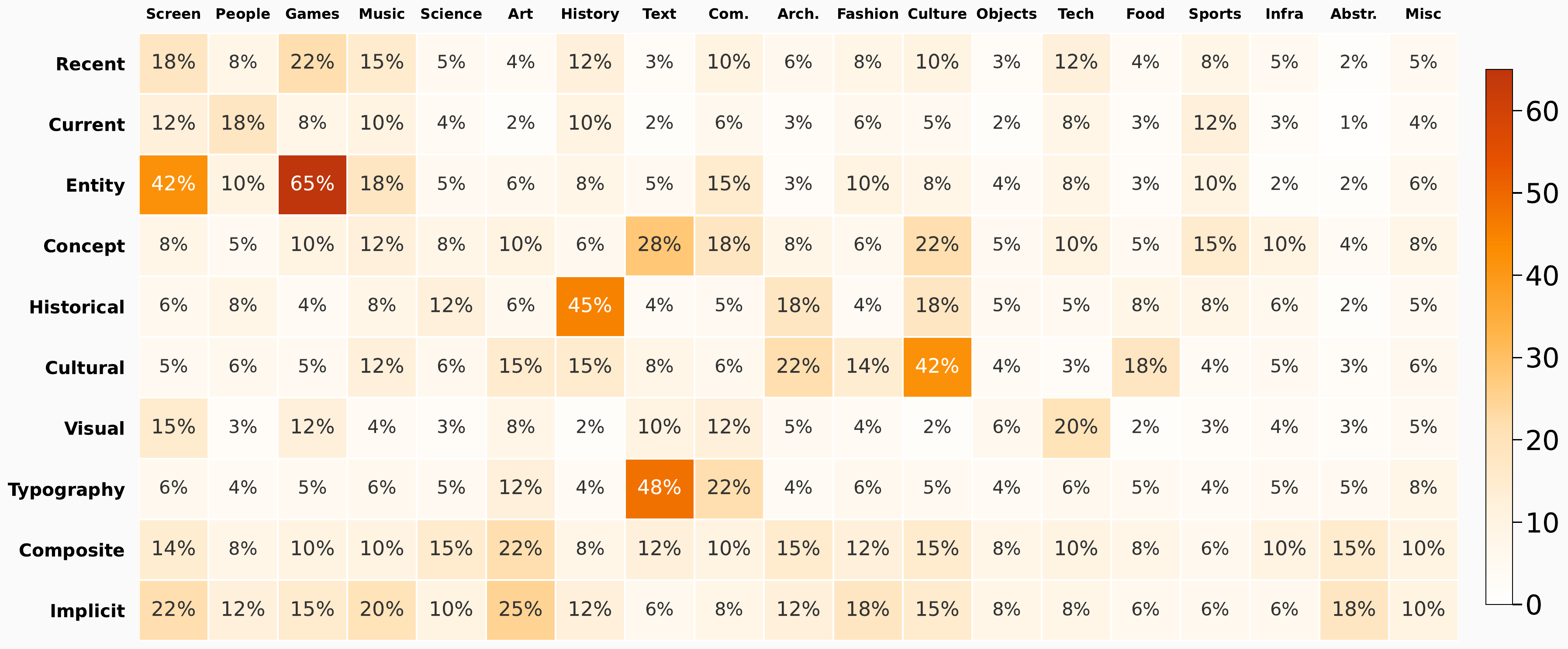}
\caption{\small\textbf{The bottleneck is pervasive: every domain--category combination shows knowledge-driven degradation.}
Heatmap summarizing cross-category structure between failure modes and domain categories in \dataset{}. The uniform severity across all cells rules out the hypothesis that failures concentrate in a few niche categories.}
\label{fig:heatmap_severity}
\end{figure}

\textbf{Layer 1 (seed entities).}
From the 10{,}840 real user prompts, we extract a seed database of
31{,}537 entities spanning 22 primary domains, each annotated with canonical
name, estimated training-set frequency, ground-truth visual references, and
distinguishing attributes. The distribution is intentionally long-tailed,
mirroring real user demand.

\textbf{Layer 2 (prompt synthesis).}
From the seed database, we synthesize the \dataset{} corpus in three steps.
First, \emph{template instantiation} produces structured drafts across categories and domains by filling parameterized slots for entities, relations, and stylistic constraints distilled from the production scrape.
Second, a \emph{frontier large language model} rewrites each instantiated template into a naturalistic user-style request while preserving grounded entities, modality requirements (image versus text retrieval), and checklist-eligible facts; temperature is elevated to diversify surface form and reduce templating artifacts.
Third, \emph{quality control} removes prompts that are ambiguous under the evaluation protocol, culturally insensitive, or misaligned with the intended failure mode; annotators also verify that checklist items remain answerable from the prompt and cached search evidence alone.
Each accepted prompt includes 2--4
ground-truth search queries, a structured list of 3--8 key visual elements
with binary verification checklists, and a 1--5 evaluation rubric tailored
to the failure category. This annotation infrastructure enables fully
automated evaluation via a Gemini-3-Flash VLM judge~\cite{nanobanana2025gemini} (Spearman $\rho = 0.87$ with human
ratings; methodology in Appendix~\ref{app:eval_correlation}).

\textbf{Cultural diversity and splits.}
The benchmark spans global cultural diversity---entities from East Asia,
South Asia, the Middle East, Africa, Latin America, and beyond---because
knowledge gaps concentrate in long-tail cultural concepts and a
Western-centric benchmark would underestimate severity. For supervised
training and model selection we use train (20{,}000), validation (128), and
\textbf{test} (751) partitions, matching Section~\ref{sec:bench_construct}; remaining non-test prompts support cached
search construction, trajectory collection, and co-training rollouts.
Unlike benchmarks that test compositionality within known
concepts~\cite{genaibench,t2icompbench,dalleval}, \bench{} targets
knowledge absence.
\section{Additional empirical analysis}
\subsection{Extended failure analysis}
\label{app:failure_ext}

\textbf{Concept corruption (gating).}
The generator's conditioning architecture typically has no reliable mechanism
for gating external information based on its own confidence in its internal
representation. When a reference image (or strongly textually specified
exemplar) is provided, it is integrated with substantial weight regardless
of whether the generator's native rendering would have been more accurate
on the same entity, yielding concept corruption on prompts where internal
knowledge was already adequate.

\textbf{Reference copying (pixel-level conditioning).}
Visual conditioning operates on pixel (and low-level feature) statistics rather
than explicit semantic attributes. Without an integration layer that specifies
which attributes to transplant, the generator tends to reproduce incidental
structure from the reference---background, lighting, framing---so the output
resembles a filtered search result rather than a new composite generation.

\textbf{Why surface cues are a poor proxy for Qwen-Image-boundary strata.}
Named entities and other shallow triggers are only loosely correlated with
the knowledge boundary inferred from Qwen-Image in Table~\ref{tab:bottleneck_overall}: many
prompts contain proper nouns that already lie inside Qwen-Image's \emph{NoSearch}
stratum, while some high-gap prompts yield few detectable entities. Cheap
gates therefore trigger search on the wrong subset, repeating the same
tradeoff between \emph{VisualSearch}/\emph{TextualSearch} gains and
\emph{NoSearch} regressions that \emph{BlindSearch} exhibits outright.

\textbf{TextualSearch stratum (extended discussion).}
The main paper (\S\ref{sec:exp_strata}) compresses the TextualSearch finding: for Klein-4B-DPO, both \CondGS and \CondGAS score below the no-search baseline because internalized contextual knowledge makes search net harmful on the hardest prompts.
Those prompts disproportionately require compositional contextual reasoning---typography, cultural composition, and implicit interpretation---that resists both single-pass parameterization and single-pass search, motivating richer co-training cycles and search strategies.

\subsection{Judge--human correlation and judge--reasoner independence}
\label{app:eval_correlation}

\textbf{Human study protocol.}
Human ratings are collected on the same 500 prompt--image pairs used to report aggregate agreement in \S\ref{sec:bench_construct}.
Each pair is presented with the user prompt, the checklist, and the category rubric; raters assign checklist judgments, rubric scores, and holistic quality judgments using the same anchors as the automated judge.
Multiple raters participate with adjudication on high-disagreement items so that gold labels are stable under resampling.

\textbf{Aggregate agreement.}
The Gemini-3-Flash judge achieves Spearman $\rho = 0.87$ with the consolidated human labels on this set, which is consistent with scalable VLM-as-judge practice for text-to-image faithfulness evaluation~\cite{hu2023tifa,wiles2024gecko,lee2024heim}.

\textbf{Stratified difficulty.}
Agreement is tightest on object-centric \emph{NoSearch} prompts where errors are predominantly omissions of salient attributes.
Agreement weakens on \emph{TextualSearch}, where glyph-level correctness and fine layout dominate rubric variance; this pattern aligns with known limitations of VLMs on dense in-image text and motivates treating TextualSearch as a conservative stratum in automated evaluation.

\section{Extended quantitative results}
\label{app:extended_quant}

\subsection{Full nine-component breakdown by stratum}
\label{app:full_component_breakdown}

Table~\ref{tab:bottleneck_full} provides the complete nine-component evaluation for all generators on both the NoSearch and Search-Intensive strata.
This supplements the six-component summary in the main paper (Table~\ref{tab:bottleneck_overall}) with the full scoring breakdown.

\begin{table}[t]
\centering
\caption{\textbf{The Search Bottleneck on \bench{} -- Full Nine-Component Breakdown.}
Scores on a 0--100 scale (one decimal); higher is better.
Every generator drops sharply on the \emph{Search-Intensive} set, confirming that the bottleneck reflects missing knowledge rather than rendering ability.
Knowledge-sensitive components (Checklist, Rubric, Visual Ref.) vary per prompt and measure knowledge presence; knowledge-invariant components (Text Rend., Phys.\ Plaus., Img.\ Qual.) measure rendering competence.}
\label{tab:bottleneck_full}
\small
\setlength{\tabcolsep}{2.0pt}
\resizebox{\columnwidth}{!}{%
\begin{tabular}{@{}llcccccccccc@{}}
\toprule
\textbf{Type} & \textbf{Generator}
  & \textbf{Overall} & \textbf{Checklist} & \textbf{Rubric} & \textbf{Prompt}
  & \textbf{Img.\ Qual.} & \textbf{Text Rend.} & \textbf{AI Nat.} & \textbf{Comp.\ Aes.}
  & \textbf{Phys.\ Plaus.} & \textbf{Vis.\ Ref.} \\
\midrule
\multicolumn{12}{@{}l}{\emph{NoSearch set (parametric knowledge sufficient)}} \\
\addlinespace[2pt]
\multirow{4}{*}{Open}
& Bagel              & 49.3 & 52.3 & 49.1 & 39.2 & 58.0 & 18.4 & 48.3 & 65.3 & 60.4 & 34.8 \\
& Flux.2-Klein-4B    & 51.0 & 54.7 & 51.8 & 39.3 & 61.3 & 13.2 & 49.5 & 67.2 & 63.5 & 32.6 \\
& Flux.2-Klein-9B    & 57.8 & 63.8 & 59.8 & 51.3 & 63.8 & 31.3 & 52.5 & 72.3 & 66.7 & 43.7 \\
& Qwen-Image         & 67.4 & 74.8 & 70.8 & 62.8 & 68.3 & 63.0 & 61.0 & 76.8 & 73.8 & 56.7 \\
\midrule
\multirow{6}{*}{Commercial}
& Qwen-Image-2       & 70.7 & 78.7 & 73.8 & 68.4 & 70.2 & 71.7 & 60.8 & 80.3 & 75.6 & 60.0 \\
& SeedDream-4.0      & 67.9 & 74.9 & 70.6 & 61.7 & 69.5 & 71.1 & 59.7 & 80.3 & 73.9 & 56.7 \\
& Nano Banana        & 63.1 & 71.1 & 67.5 & 59.7 & 66.3 & 42.8 & 56.5 & 76.3 & 68.5 & 53.8 \\
& Nano Banana Pro    & 75.0 & 82.8 & 78.1 & 72.8 & 71.5 & 85.9 & 65.0 & 83.3 & 78.2 & 67.8 \\
& GPT-Image-2        & 71.1 & 78.6 & 75.6 & 70.8 & 69.0 & 67.7 & 60.8 & 77.7 & 73.9 & 64.3 \\
\midrule
\multicolumn{12}{@{}l}{\emph{Search-Intensive (external knowledge required)}} \\
\addlinespace[2pt]
\multirow{4}{*}{Open}
& Bagel              & 21.5 & 18.2 & 17.6 & 13.3 & 30.5 &  2.5 & 29.3 & 33.6 & 36.8 & 13.5 \\
& Flux.2-Klein-4B    & 24.1 & 19.8 & 18.4 & 12.4 & 37.2 &  4.2 & 33.6 & 39.3 & 46.2 & 11.9 \\
& Flux.2-Klein-9B    & 26.7 & 24.2 & 23.1 & 17.2 & 36.8 &  7.2 & 32.9 & 40.4 & 48.6 & 16.9 \\
& Qwen-Image         & 27.9 & 24.8 & 24.3 & 18.6 & 40.1 &  8.7 & 31.6 & 42.8 & 44.6 & 17.7 \\
\midrule
\multirow{6}{*}{Commercial}
& Qwen-Image-2       & 31.6 & 28.5 & 27.1 & 22.1 & 42.2 & 12.7 & 36.3 & 45.5 & 48.2 & 21.0 \\
& Imagen3-Fast       & 14.1 &  9.7 & 10.0 &  6.9 & 22.2 &  1.4 & 21.4 & 23.2 & 23.7 &  7.0 \\
& SeedDream-4.0      & 45.9 & 44.2 & 43.6 & 38.5 & 57.0 & 35.9 & 47.1 & 58.7 & 64.0 & 35.1 \\
& Nano Banana        & 44.1 & 41.0 & 40.4 & 36.0 & 57.1 & 28.0 & 47.4 & 61.5 & 65.5 & 33.2 \\
& Nano Banana Pro    & 65.3 & 64.4 & 63.1 & 60.7 & 71.4 & 65.0 & 62.0 & 75.9 & 78.5 & 58.3 \\
& GPT-Image-2        & 71.0 & 71.2 & 70.1 & 69.2 & 75.1 & 75.9 & 64.7 & 80.4 & 77.3 & 66.0 \\
\bottomrule
\end{tabular}%
}
\end{table}

\section{Protocols and supplementary artifacts}
\subsection{End-to-end workflow and reasoner I/O}
\label{app:prompts}
\label{app:reasoner_io}

The reasoner's three stages of \S\ref{sec:atomic}---\textbf{gate} (Stage~1),
\textbf{filter} (Stage~2), and \textbf{integrate} (Stage~3)---are stored in our SFT
corpus as trajectory types Task~A, Task~B, and Task~C, respectively.
We first trace a \emph{single} request end-to-end to show the overall inputs and outputs,
then give one representative input--output pair per stage so the input contract and output
fields (shown as \texttt{field}) are visible at a glance. We render outputs in compact
form; the released co-training corpus contains the verbatim JSON in the exact schema.
These are representative \emph{successful} cases; failure modes such as copy effects are
analyzed in \S\ref{sec:naive}.

\begin{figure}[t]
  \centering
  \setlength{\tabcolsep}{3pt}
  \begin{tabular}{@{}ccc@{}}
    \includegraphics[width=0.30\linewidth]{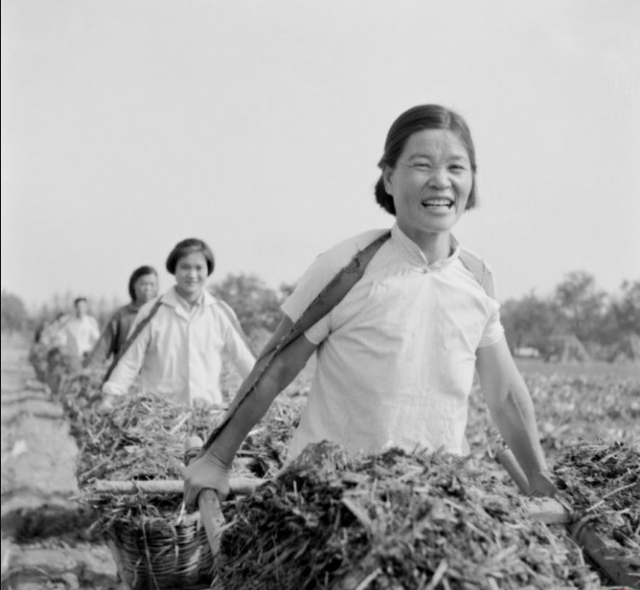} &
    \includegraphics[width=0.30\linewidth]{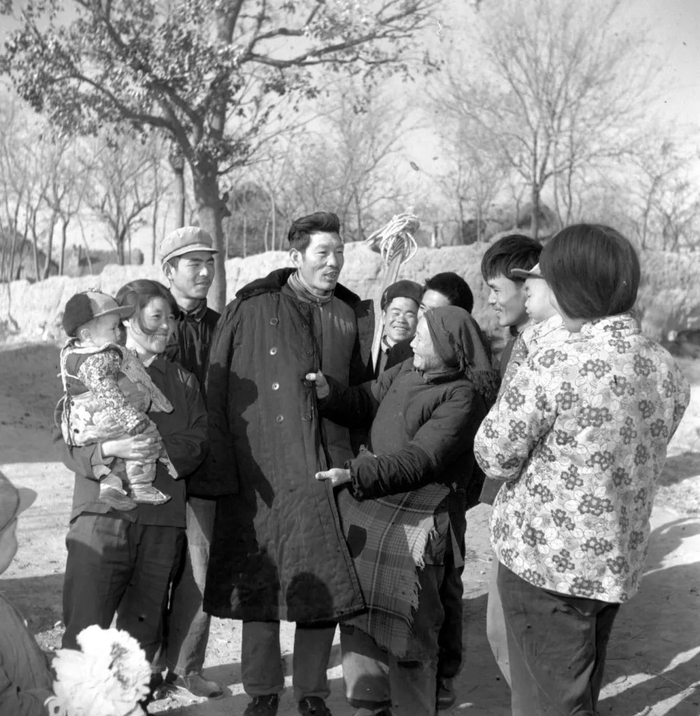} &
    \includegraphics[width=0.30\linewidth]{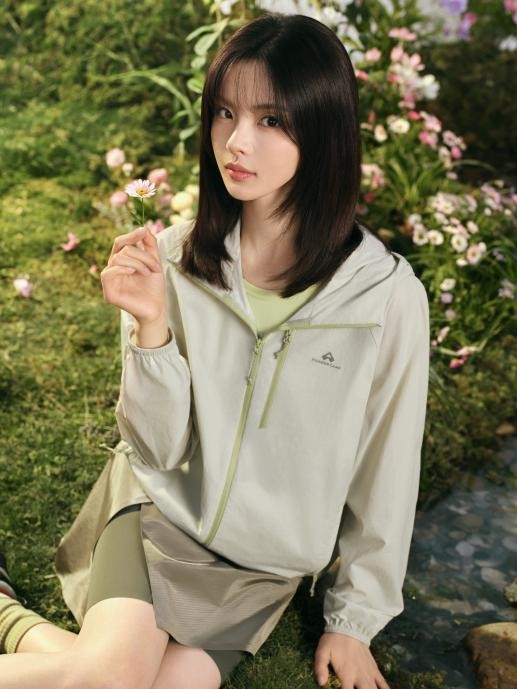} \\
    \scriptsize Reference~1 (scene) & \scriptsize Reference~2 (costume) & \scriptsize Reference~3 (likeness)
  \end{tabular}\\[6pt]
  \includegraphics[width=0.55\linewidth]{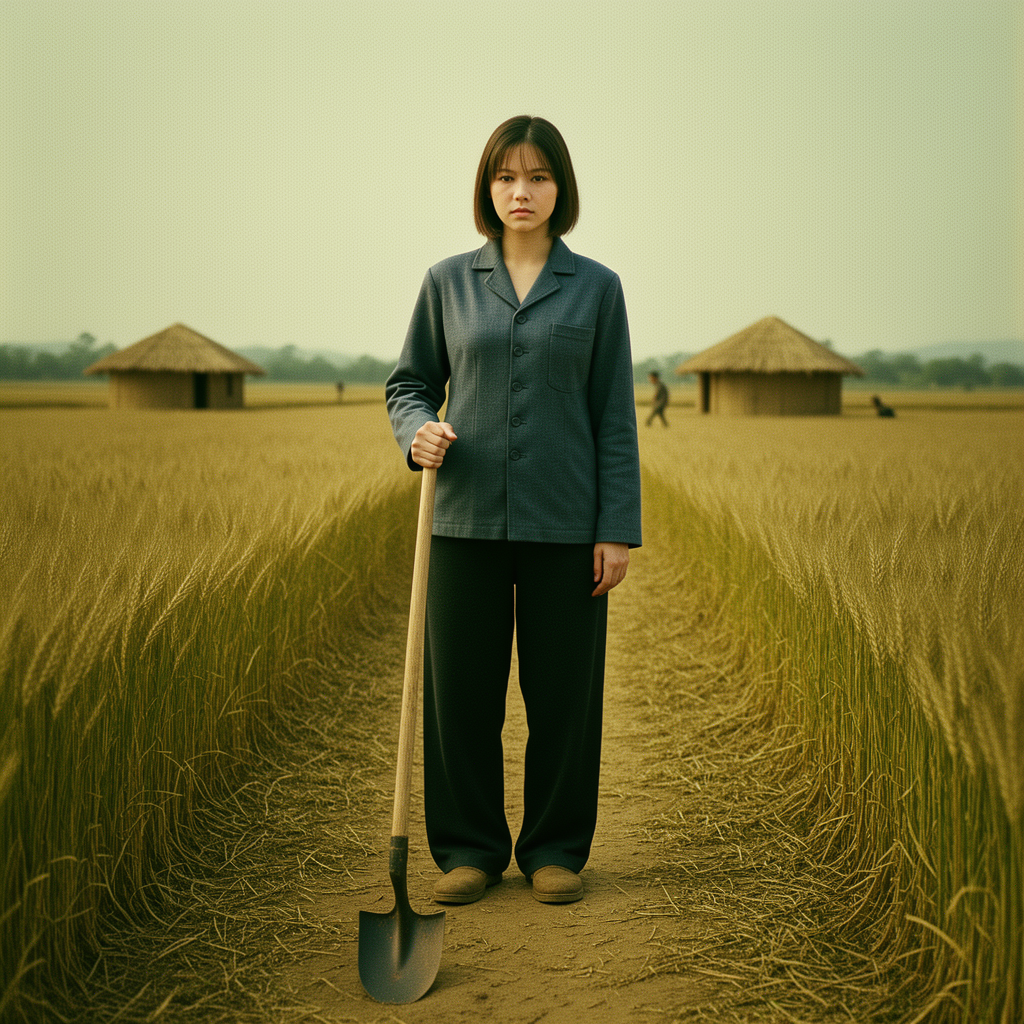}\\[2pt]
  {\scriptsize Augmented output (Flux.2-Klein-4B-DPO).}
  \caption{\textbf{The end-to-end example's visual output.} The three references the reasoner
  retrieved and selected (top: scene, costume, likeness) and the image the generator produced
  from the enriched prompt (bottom), for the Yang Chaoyue request traced in the box above.
  This is the section's only image example; it grounds what ``retrieved references'' and
  ``generation'' concretely look like. The per-stage boxes that follow are text-only and use
  different prompts, each chosen to isolate one stage's behavior.}
  \label{fig:workflow_overview}
\end{figure}

\paragraph{End-to-end workflow (condensed).}
A user prompt enters; the reasoner runs gate\,$\to$\,filter\,$\to$\,integrate to decide
what to search, select references, and emit an enriched prompt; the generator renders the
image; and a VLM judge scores it against a checklist and rubric. The box below traces one
request through these steps, and Figure~\ref{fig:workflow_overview} shows the retrieved
references and the resulting generation.

\begin{examplebox}
\small
\textbf{End-to-end example (condensed).}\\[2pt]
\textbf{Input --- user request:} \emph{``Paint Yang Chaoyue on a 1970s Chinese rural field ridge; commune-member clothing; long-handled hoe; earth tones; 1970s film grain.''}\\[3pt]
\textbf{Stage~1 (gate) --- image search queries:} Yang Chaoyue natural photo; 1970s commune labor scene; 1970s rural film photography; long-handled hoe.\\[2pt]
\textbf{Stage~2 (filter) --- selected references:} a period rural-ridge scene; commune work clothing; Yang Chaoyue likeness.\\[2pt]
\textbf{Stage~3 (integrate) --- enriched prompt (output to generator):} \emph{``A 1970s Chinese rural labor scene on a yellow-earth field ridge: Yang Chaoyue in period commune clothing, holding a wooden long-handled hoe with a curved iron blade, natural expression, earth-tone palette, slight vintage color cast, film grain, no modern elements.''}\\[2pt]
\textbf{Output --- generation \& evaluation:} the enriched prompt and the three references condition the generator (Figure~\ref{fig:workflow_overview}); a VLM judge then scores the result on likeness, historical authenticity, and atmospheric tone (protocol in \S\ref{app:eval}).
\end{examplebox}

\paragraph{Stage~1 --- Gate (SFT Task~A).}
The gate decides \emph{whether} external grounding is needed, classifies each surviving
gap by type and severity, and proposes modality-labeled search queries. The two cases
below show the decision it must make: it \emph{fires} on a named landmark with a critical
visual-identity gap, and it \emph{skips} a generic studio prompt the base generator
already renders reliably---the discipline that keeps the pipeline from over-retrieving.

\begin{examplebox}
\small
\textbf{Stage~1 (gate) --- search fires.}\\[2pt]
\textbf{Prompt:} \emph{``Photorealistic dusk poster of the Shanghai Tower showing its actual spiral facade and correct height proportions.''}\\[3pt]
\textbf{Knowledge gap:} Shanghai Tower --- \texttt{category}: entity identity; \texttt{severity}: \emph{critical}. The signature 120-degree spiral taper and tiered glass facade are frequently rendered incorrectly; an image reference fixes the exact geometry.\\[2pt]
\textbf{Decision} (\texttt{search\_decision}): \emph{search}. \quad \textbf{Query} (\texttt{modality}: image): ``Shanghai Tower full exterior dusk photograph''.\\[2pt]
\textbf{Justification:} a critical visual-identity gap survives analysis, so image search is warranted.
\end{examplebox}

\begin{examplebox}
\small
\textbf{Stage~1 (gate) --- search skips.}\\[2pt]
\textbf{Prompt:} \emph{``Photorealistic product shot of a blue and white porcelain teapot with bamboo motif on a marble tabletop, shallow depth of field, soft studio lighting.''}\\[3pt]
\textbf{Knowledge gaps} (\texttt{knowledge\_gaps}): none. Porcelain teapots, bamboo motifs, marble surfaces, and studio lighting all render reliably from training data; no named entities, events, or cultural specifics are present.\\[2pt]
\textbf{Decision} (\texttt{search\_decision}): \emph{skip}. No critical or important gap survives; searching would inject noise without improving accuracy.\\[2pt]
{\footnotesize\emph{Fields:} \texttt{severity}$\in$\{critical, important, moderate, minimal\}; \texttt{modality}$\in$\{image, web\}.}
\end{examplebox}

\bigskip

\paragraph{Stage~2 --- Filter (SFT Task~B).}
Once the gate fires, retrieval returns several candidate references; the filter selects
the single one that most directly fills the identified gap while minimizing extraneous
content. Below, only one of three candidates carries the required location-specific
layout, and the filter selects it by 0-based index.

\begin{examplebox}
\small
\textbf{Stage~2 (filter) --- reference selection.}\\[2pt]
\textbf{User request:} \emph{``Create a photorealistic render of the current Atlanta, GA UV index dashboard, showing both the real-time UV reading and the daily maximum forecast value.''}\\[2pt]
\textbf{Search query:} ``Official Atlanta Georgia real-time UV index dashboard example''.\\[3pt]
\textbf{Candidates} (0-based):
\begin{itemize}[leftmargin=1.2em,itemsep=1pt,topsep=1pt,parsep=0pt]
  \item[0.] generic Home Assistant UV-index card (dashboard widget)
  \item[1.] Atlanta, GA real-time UV index dashboard
  \item[2.] Buckhead dermatology clinic UV-safety page
\end{itemize}
\vspace{1pt}
\textbf{Identified gap} (\texttt{identified\_knowledge\_gaps}): the model lacks the location-specific layout, branding, and data structure of a real-time UV dashboard for Atlanta.\\[2pt]
\textbf{Selected} (\texttt{selected\_index}): \textbf{1}. Only candidate~1 shows an Atlanta-specific dashboard with the correct layout; candidates~0 and~2 are a generic widget and a clinical page and lack the required city-specific structure.
\end{examplebox}

\bigskip

\paragraph{Stage~3 --- Integrate (SFT Task~C).}
The final stage folds the selected reference(s) into a generation-ready prompt, routing
visual grounding \emph{through language}: rather than passing raw pixels alone, the
reasoner names exactly what to borrow with a grounded citation (``following Image~1,
render~\dots'') so the generator inherits the missing knowledge without copying
unrequested style or layout.

\begin{examplebox}
\small
\textbf{Stage~3 (integrate) --- reference-grounded refinement.}\\[2pt]
\textbf{User request:} \emph{``Generate a photorealistic image of a polar bear standing in the Sahara Desert, looking confused as it searches for ice, with sand dunes and a single dead acacia tree nearby.''}\\[2pt]
\textbf{Reference~1:} a dead acacia tree under hard desert sunlight (retrieved for the tree's branching structure).\\[3pt]
\textbf{Borrow from Image~1} (\texttt{borrow\_from\_references}): the acacia tree's forked branching silhouette, dry cracked bark texture, and hard midday desert lighting.\\[2pt]
\textbf{Create new} (\texttt{create\_new}): the confused polar bear searching for ice; rolling Sahara dunes to the horizon; clear blue sky.\\[2pt]
\textbf{Refined prompt} (\texttt{refined\_prompt}): \emph{``Photorealistic 8k polar bear in the Sahara, furrowed brow as it searches for ice. \underline{Following Image~1}, render the dead acacia tree with the same forked branches, dry cracked bark, and hard midday sunlight; match that lighting on the bear so its white fur casts the same sharp shadows. Golden-orange dunes roll to the horizon under a clear blue sky.''}
\end{examplebox}

\subsection{Evaluation protocol}
\label{app:eval}

\paragraph{Automated judge protocol.}
Each evaluation instance gives the judge the user prompt, the verification checklist, and the category-specific rubric, together with any reference images carrying short textual annotations (entity focus and severity).
References are used only as comparison evidence; the single ``image to assess'' is scored.
The judge follows a fixed sequence: recover task requirements; separate reference versus scored roles; extract observable evidence (subjects, attributes, scene, style, in-image text, artifacts); compare against references where checklist items depend on them; assign anchored half-point scores for checklist rows, rubric dimensions, and generic quality axes; then emit scored blocks for reference alignment, checklist, rubric, visual-reference fidelity, and text-reference fidelity, suitable for aggregation into the nine reported components (Table~\ref{tab:judge_components}).

\bigskip

\paragraph{Score components and aggregation.}
All scores use a $[0,3]$ scale in $0.5$ steps ($0$~=~not met/contradicted, $1$~=~weak with major issues, $2$~=~mostly met with minor issues, $3$~=~fully met).
Each score is accompanied by an auditable reason in three parts: \emph{Anchor} (initial score with evidence), \emph{Adjustments} (signed deltas with justification), and \emph{Computed final} (clamped sum matching the emitted score).
Table~\ref{tab:judge_components} lists the nine reported components and their judge instructions.
The overall score is the mean of present components, rescaled to $0$--$100$.

\begin{table}[h]
\centering
\caption{\textbf{Judge scoring components.} Each component is scored on $[0,3]$. Components 1--2 are prompt-specific; components 3--8 are fixed generic dimensions scored for every image; component 9 is conditional on whether reference images are provided.}
\label{tab:judge_components}
\small
\renewcommand{\arraystretch}{1.25}
\begin{tabularx}{\linewidth}{@{}clX@{}}
\toprule
\textbf{\#} & \textbf{Component} & \textbf{Judge instruction (abbreviated)} \\
\midrule
1 & Checklist & Mean of per-item verification scores. Each prompt carries 3--10 binary-verifiable items (e.g., ``Is the Om symbol correctly formed?''); the judge scores each against observable evidence. \\
2 & Rubric adaptive & Weighted mean of prompt-specific rubric dimensions (e.g., \emph{symbol\_accuracy}, \emph{historical\_fidelity}) excluding the six fixed generic dimensions below. \\
\midrule
3 & Prompt faithfulness & Are all requested subjects, attributes, actions, scene, and style present and accurate? \\
4 & Image quality & Visual clarity and sharpness; absence of artifacts or defects; style, lighting, and perspective coherence. \\
5 & Text rendering & Does rendered in-image text match required content? Readable, correctly spelled, well-placed? Score left empty if no text is required and none is present. \\
6 & AI naturalness & Organic texture realism vs.\ AI smoothness; fine-detail plausibility vs.\ uncanny uniformity; environment grounded vs.\ dreamlike or generic. \\
7 & Composition \& aesthetics & Framing and balance; depth and spatial arrangement; color harmony and consistent lighting; overall visual appeal. \\
8 & Physical plausibility & Anatomy (fingers, limbs, proportions, pose); object physics (gravity, support, balance); spatial consistency (perspective, scale, occlusion); material properties; lighting and shadow consistency. Score left empty for abstract or stylized content. \\
\midrule
9 & Visual reference fidelity & Scored only when reference images are attached. Identity preservation (facial features, distinguishing traits); attribute consistency (colors, textures, proportions); style fidelity. Skipped (not zero) when no references are provided. \\
\bottomrule
\end{tabularx}
\end{table}

\paragraph{Difficulty partition (Set~I--III).}
The 651 search-intensive prompts are partitioned into three difficulty sets using Nano Banana Pro no-search quality as the sorting variable: each prompt receives an overall score from Nano Banana Pro \emph{without any retrieval augmentation}, and tercile boundaries on this score distribution define Set~I (easiest third), Set~II, and Set~III (hardest third).
The partition is fixed across all experiments; the 100-prompt NoSearch subset is reported as a separate column.

\paragraph{Judge fidelity.}
The judge achieves Spearman $\rho = 0.87$ with human ratings on a
held-out set of 500 prompt--image pairs
(Appendix~\ref{app:eval_correlation}). The Anchor/Adjustment/Computed
diagnostic encourages falsifiable critiques of individual scores.

\section{Hardware and implementation details}
\label{app:hardware}
All experiments were conducted on NVIDIA H20 (80~GB HBM3) GPUs.
Below we detail the compute configuration for each training phase.

\subsection{VLM reasoner training (Phase~0 and Phase~2)}
The VLM reasoner (Qwen3-VL-8B) is trained using the ModelScope-Swift framework (\texttt{ms-swift}).
We perform full-parameter supervised finetuning with the configuration in Table~\ref{tab:app_reasoner_sft}.

\begin{table}[t]
  \centering
  \caption{\textbf{VLM reasoner SFT hyperparameters} (Phase~0 and Phase~2).}
  \label{tab:app_reasoner_sft}
  \small
  \begin{tabularx}{\linewidth}{lX}
    \toprule
    \textbf{Hyperparameter} & \textbf{Value} \\
    \midrule
    Base model & Qwen3-VL-8B-Instruct \\
    Training type & Full-parameter SFT \\
    Precision & bfloat16 \\
    Epochs & 5--6 \\
    Learning rate & $1\times 10^{-5}$ \\
    Warmup ratio & 0.05 \\
    Per-device batch size & 1 \\
    Gradient accumulation steps & 8 \\
    Effective batch size & 64 ($8$ GPUs $\times 1 \times 8$ accum.) \\
    Max sequence length & 12{,}288 tokens \\
    Max image pixels & 262{,}144 ($512^2$) \\
    Image max token number & 512 \\
    Attention implementation & FlashAttention-2 \\
    Sequence packing & Enabled (padding-free) \\
    Gradient checkpointing & Enabled \\
    ViT parameters & Frozen \\
    DeepSpeed stage & ZeRO-2 \\
    Hardware & $8 \times$ NVIDIA H20 (80~GB) \\
    Training data & ${\sim}20$K SFT trajectories \\
    Wall-clock time & ${<}4$~hours \\
    \bottomrule
  \end{tabularx}
\end{table}

The SFT dataset consists of approximately 20{,}000 expert-annotated reasoning trajectories for the gate--filter--compress pipeline (Tasks A/B/C), stored in the training-row format consumed by ModelScope-Swift.
A 1\% held-out split is used for validation.
We use dataset preprocessing with 200 workers and 4 dataloader workers for efficient data loading.

\subsection{Generator DPO training (Phase~1)}
Online iterative DPO for the visual generator (Flux.2-Klein-4B) is conducted using the Flow-Factory framework with the configuration in Table~\ref{tab:app_dpo}.

\begin{table}[t]
  \centering
  \caption{\textbf{Generator DPO hyperparameters} (Phase~1).}
  \label{tab:app_dpo}
  \small
  \begin{tabularx}{\linewidth}{lX}
    \toprule
    \textbf{Hyperparameter} & \textbf{Value} \\
    \midrule
    Base model & Flux.2-Klein-4B \\
    Finetuning type & LoRA \\
    LoRA rank / alpha & 64 / 128 \\
    Target modules & Default (attention layers) \\
    Precision & bfloat16 \\
    Learning rate & $1\times 10^{-4}$ \\
    Optimizer & AdamW ($\beta_1{=}0.9$, $\beta_2{=}0.999$, $\varepsilon{=}10^{-8}$) \\
    Weight decay & $1\times 10^{-4}$ \\
    Max gradient norm & 1.0 \\
    Resolution & $512 \times 512$ \\
    Condition image size & $512 \times 512$ \\
    Per-device batch size & 1 \\
    Group size (candidates per prompt) & 5 \\
    Unique samples per epoch & 64 \\
    Gradient steps per epoch & 2 \\
    Gradient accumulation & Auto (computed from group size) \\
    DPO $\beta$ & 100.0 \\
    Reference model device & CPU (offloaded) \\
    Advantage aggregation & Group-relative DPO \\
    EMA decay / update interval & 0.99 / 4 steps \\
    Guidance scale & 4.0 \\
    Inference steps (sampling) & 50 \\
    Timestep sampling & Logit-normal ($\mu{=}0.0$, $\sigma{=}1.0$) \\
    Gradient checkpointing & Enabled \\
    Hardware & $8 \times$ NVIDIA H20 (80~GB) \\
    Training data & ${\sim}7$K prompt--image preference pairs \\
    Save frequency & Every 10 gradient steps \\
    \bottomrule
  \end{tabularx}
\end{table}

The DPO reward signal is provided by the Qwen3-VL-8B VLM judge, served via vLLM on a separate node (accessed at inference time over HTTP).
The judge evaluates generated candidates using the \dataset{} evaluation protocol with nine scoring dimensions.
Judge inference uses a maximum of 4{,}096 output tokens, temperature 0.2, and a maximum pixel budget of 262{,}144 per image, with up to 24 concurrent requests.

\paragraph{Preference construction.}
For each prompt, the generator produces 5 candidate images (one additional candidate beyond the group for preference diversity via \texttt{preference\_extra\_candidates: 1}).
Candidates are scored by the VLM judge, and preference pairs are constructed from the highest- and lowest-scored generations within each group.
A structural similarity (SSIM) penalty (\texttt{dpo\_penalize\_condition\_ssim: 0.95}) discourages degenerate pairs where chosen and rejected images are near-identical.

\paragraph{Flow-matching dynamics.}
We use a Flow-SDE scheduler with noise level 1.0, time shift 1.0, and timestep range truncated at 0.1 for stable training.

\subsection{Reasoner rejection finetuning (Phase~2)}
Phase~2 recalibrates the reasoner to the DPO-strengthened generator using rejection-sampling finetuning.
The training infrastructure is identical to Phase~0 (ModelScope-Swift, $8 \times$ H20, same hyperparameters), with the key difference that the training data consists of filtered trajectories: only reasoner rollouts that produce positive group-relative advantage (i.e., search improved the strengthened generator's output quality) are retained for training.
The full RFT cycle completes in approximately $4 \times 8 = 32$ GPU-hours.

\subsection{Total compute budget}
Table~\ref{tab:app_compute_budget} summarizes end-to-end training compute (Phase~1 GPU-hours are $8$ GPUs $\times 24$~h wall-clock).
The VLM judge service runs concurrently on a separate GPU and is not folded into the training GPU-hour totals.

\begin{table}[t]
  \centering
  \caption{\textbf{Compute budget overview.} Training GPU-hours are approximate.}
  \label{tab:app_compute_budget}
  \small
  \begin{tabular}{@{}lccc@{}}
    \toprule
    \textbf{Phase} & \textbf{GPUs} & \textbf{Wall-clock} & \textbf{GPU-hours} \\
    \midrule
    Phase~0: Reasoner SFT & $8 \times$ H20 & ${<}4$~h & ${\sim}32$ \\
    Phase~1: Generator DPO & $8 \times$ H20 & 24~h & ${\sim}192$ \\
    Phase~2: Reasoner RFT & $8 \times$ H20 & ${\sim}4$~h & ${\sim}32$ \\
    VLM Judge serving (vLLM) & $1 \times$ H20 & concurrent & --- \\
    \midrule
    \textbf{Total (training)} & --- & --- & \textbf{${\sim}256$} \\
    \bottomrule
  \end{tabular}
\end{table}

All training runs use CUDA memory optimization via \\
\texttt{PYTORCH\_CUDA\_ALLOC\_CONF=expandable\_segments:True}.
Distributed training communication uses NCCL over NVLink interconnects within each 8-GPU node.

\section{Limitations and future directions}
\label{app:limitations}
Our work deliberately prioritizes depth of insight over exhaustive optimization, and this choice opens several promising avenues for future research.

\paragraph{Iterative co-training.}
Our recipe consists of a single DPO pass followed by a single RFT pass.
While this minimal protocol already produces monotonic improvement and validates the core principle, the knowledge boundary between internalizable and contextual knowledge is unlikely to be fully resolved in one iteration.
Multi-round co-training---where the generator and reasoner alternate improvement over several cycles---could progressively sharpen this boundary and unlock additional gains.
Investigating convergence behavior, diminishing returns, and potential instability in such iterative loops is an exciting direction.

\paragraph{Generator scale and architecture.}
Our co-training experiments use a 4B-parameter generator (Flux.2-Klein-4B) and 7B-parameter generator (Bagel).
The substantial gap between our best result and frontier commercial systems suggests that larger or architecturally different generators may internalize a broader swath of knowledge, shifting the knowledge boundary and potentially changing the distribution of prompts that require search.
Studying how the internalizable/contextual split evolves across generator scales---and whether certain failure categories (e.g., temporal-recent vs.\ cultural-specificity) exhibit qualitatively different scaling behavior---would deepen understanding of the knowledge boundary as a function of model capacity.

\paragraph{Reward signals.}
Our current evaluation relies on an automated VLM judge (Gemini-3-Flash).
While it achieves strong correlation with human ratings (Spearman $\rho = 0.87$), such judges can introduce noisy rewards---a widely acknowledged issue when automated scoring substitutes for human judgment~\cite{wiles2024gecko,lee2024heim}.
The choice of DPO is intentional: it resists reward noise by selecting the best- and worst-ranked generations to construct preference pairs~\cite{rafailov2023dpo}.

\section{Broader impact}
\label{app:broader}
This work contributes positively to the research community and society in several important ways, while also carrying risks that warrant consideration.

\paragraph{Positive impacts.}
\begin{itemize}
  \item \textbf{New evaluation resources for an underserved problem.}
  \dataset{} and \bench{} fill a significant evaluation gap.
  Existing text-to-image benchmarks predominantly test prompts that fall within generators' parametric knowledge, creating an illusion of near-solved performance.
  By systematically cataloging twelve failure categories grounded in over 10{,}000 real user prompts, our benchmark surfaces a class of failures that affects daily production use but has been largely invisible to the research community.
  Making this gap measurable is a prerequisite for closing it.

  \item \textbf{A long-tailed, culturally diverse, and challenging reference-to-image dataset.}
  The released corpus of 20{,}000 prompts, spanning twenty-two domains and intentionally representing global cultural diversity---including underrepresented cultural symbols, regional artifacts, non-Latin typography, and entities from the Global South---provides a uniquely challenging testbed.
  The long-tailed entity distribution mirrors real user demand and offers the community a resource for studying knowledge-intensive generation at a scale and diversity not previously available.
  The accompanying pre-executed search results (both image and web returns) form a fully offline search environment, lowering the barrier for reproducible research on search-augmented visual generation without requiring live API access or incurring search costs.

  \item \textbf{Advancing equitable visual generation.}
  A disproportionate share of search-intensive failures affects culturally specific content: regional festivals, indigenous art forms, historical figures outside Western canons, and scripts beyond Latin.
  By explicitly benchmarking these categories and demonstrating that search-augmented generation can improve fidelity for such content, this work contributes toward more equitable visual AI that can serve diverse global users rather than only those whose cultural knowledge is well-represented in training data.

  \item \textbf{Computational accessibility.}
  The deliberately minimal co-training recipe (one DPO pass, one RFT pass, a 4B generator, an 8B reasoner, fitting in $4\times 8$ GPU hours for the RFT phase) demonstrates that meaningful progress on knowledge-intensive generation does not require frontier-scale compute, broadening access to researchers with limited resources.
\end{itemize}

\paragraph{Potential negative impacts and mitigations.}
\begin{itemize}
  \item \textbf{Misinformation risk.}
  Improving generators' ability to render specific real-world entities---public figures, cultural symbols, historical events---could lower the barrier for producing convincing visual misinformation.
  We note, however, that our benchmark explicitly includes an ``AIGC fakeness'' evaluation dimension that measures detectable artifacts, and our released assets are evaluation prompts and search results rather than trained generator checkpoints.
  We encourage downstream users to pair improved generation with provenance tracking and watermarking.

  \item \textbf{Privacy concerns.}
  Some search-intensive prompts involve public figures.
  Our dataset construction follows standard practices: all seed entities are drawn from publicly available information, and the benchmark is designed for research evaluation rather than generating deceptive impersonations.
  We will include usage guidelines with the released dataset that prohibit malicious identity synthesis.

  \item \textbf{Bias amplification.}
  While we intentionally design for cultural diversity, the search results that form part of our released offline environment inherit biases from web search engines.
  Researchers using these results should be aware that search-returned references may reflect stereotypical or incomplete representations of certain cultures, and should evaluate whether such biases propagate into generated outputs.
  {\color{white}\nocite{wang2025code,wang2025emergent,wang2025pixelreasoner,wang2025vl,wang2025reverse,wang2026badseeingbadthinking}}
\end{itemize}

\subsection{Third-party models, data access, and terms of use}
\label{app:third_party}
\textbf{Open-weight generators.}
Bagel~\cite{deng2025bagel}; FLUX.2 Klein checkpoints~\cite{blackforest2026flux2}; Qwen-Image~\cite{qwenimage}.
\textbf{Commercial APIs.}
Qwen-Image-2~\cite{qwenimage}; Gemini developer endpoints (Gemini~2.5~Flash Image, Gemini~3~Pro Image; consumer names Nano Banana / Nano Banana Pro in some regions)~\cite{nanobanana2025gemini}; ByteDance Jimeng~\cite{jimeng2025bytedance}; SeedDream 4.0~\cite{seedream2025arxiv}; Imagen 3 Fast~\cite{baldridge2024imagen3}.
Public documentation does not specify whether these endpoints internally invoke retrieval, tool use, or proprietary knowledge bases at inference time.
Consequently, high \bench{} scores for commercial systems should be interpreted as end-to-end product behavior rather than as pure comparisons to open-weight generators under identical conditioning assumptions.

\textbf{Use.}
Experiments respect each provider's publicly posted terms and documentation; releases from this work consist of prompts and cached search artifacts, not redistributed model weights or proprietary API payloads.

\subsection{Responsible release and safeguards}
\label{app:responsible_release}

\textbf{Release scope.}
\dataset{} and \bench{} are intended for research evaluation of knowledge-intensive visual generation.
The public release will include prompts, annotations, evaluation checklists, rubrics, search queries, and permissible search metadata or derived attributes, rather than raw scraped images or proprietary web content.
We do not redistribute copyrighted source images, proprietary web content, or proprietary API payloads without permission.

\textbf{Safeguards and intended use.}
To reduce copyright and safety risks, we filter or exclude unsafe search artifacts, including explicit sexual content, graphic violence, hateful or extremist material, and personally sensitive content.
The benchmark is intended for evaluating search-intensive visual generation, not for producing deceptive, infringing, or harmful content.
Users of the released assets should respect the licenses and terms of the original sources and should not use the benchmark to generate unauthorized reproductions of protected characters, brands, or individuals.


\end{document}